%% file: main.tex
\documentclass[10pt, twocolumn, letterpaper]{article}

\usepackage[english]{babel}
\usepackage[utf8]{inputenc}
\usepackage{graphicx} % Required for inserting images
\usepackage{amsmath}
\usepackage{amssymb}
\usepackage{bm}
\usepackage{caption} 
\usepackage{multirow}
\usepackage{xcolor}
\usepackage{booktabs}
\PassOptionsToPackage{hyphens}{url}
\usepackage{setspace}
\usepackage[super,sort&compress]{natbib}

\usepackage[pagebackref,breaklinks,colorlinks,citecolor=blue]{hyperref}
\usepackage[capitalize]{cleveref}
\crefname{section}{Sec.}{Secs.}
\Crefname{section}{Section}{Sections}
\Crefname{table}{Table}{Tables}
\crefname{table}{Tab.}{Tabs.}
\Crefname{figure}{Figure}{Figures}
\crefname{figure}{Fig.}{Figs.}

\title{Deep Unsupervised Anomaly Detection in Brain Imaging: Large-Scale Benchmarking and Bias Analysis}
\author{Alexander Frotscher$^a$, Christian F. Baumgartner$^{b,e}$, Thomas Wolfers$^{a,c,d}$}
\date{%
    $^a$\small Department of Psychiatry and Psychotherapy, University Hospital Tübingen, Tübingen, Germany\\%
    $^b$\small Cluster of Excellence – Machine Learning for Science, University of Tübingen, Baden-Württemberg, Germany\\%
    $^c$\small Department of Psychology, Friedrich Schiller University of Jena, Germany\\%
    $^d$\small German Center for Mental Health (DZPG), partner site Halle/Jena/Magdeburg, Germany\\%
    $^e$\small University of Lucerne, Lucerne, Switzerland\\ [2ex]%
    \today
}
\begin{document}

\maketitle

\begin{abstract}
Deep unsupervised anomaly detection in brain magnetic resonance imaging offers a promising route to identify pathological deviations without requiring lesion-specific annotations. Yet, fragmented evaluations, heterogeneous datasets, and inconsistent metrics have hindered progress toward clinical translation. Here, we present a large-scale, multi-center benchmark of deep unsupervised anomaly detection for brain imaging. The training cohort comprised 2,976 T1 and 2,972 T2-weighted scans ($\approx$\,461,000 slices) from healthy individuals across six scanners, with ages ranging from 6 to 89 years. Validation used 92 scans to tune hyperparameters and estimate unbiased thresholds. Testing encompassed 2,221 T1w and 1,262 T2w scans spanning healthy datasets and diverse clinical cohorts. Across all algorithms, the Dice-based segmentation performance varied between $\approx$\,0.03 and $\approx$\,0.65, indicating substantial variability and underscoring that no single method achieved consistent superiority across lesion types or modalities for any task. To assess robustness, we systematically evaluated the impact of different scanners, lesion types and sizes, as well as demographics (age, sex). \textit{Reconstruction-based} methods, particularly diffusion-inspired approaches, achieved the strongest lesion segmentation performance, while \textit{feature-based} methods showed greater robustness under distributional shifts. However, systematic biases, such as scanner-related effects, were observed for the majority of algorithms, including that small and low-contrast lesions were missed more often, and that false positives varied with age and sex. Increasing healthy training data yields only modest gains, underscoring that current unsupervised anomaly detection frameworks are limited algorithmically rather than by data availability. Our benchmark establishes a transparent foundation for future research and highlights priorities for clinical translation, including image native pretraining, principled deviation measures, fairness-aware modeling, and robust domain adaptation.
\end{abstract}

\section{Introduction}
\begin{figure*}[ht]
    \centering
    \includegraphics[scale=0.8]{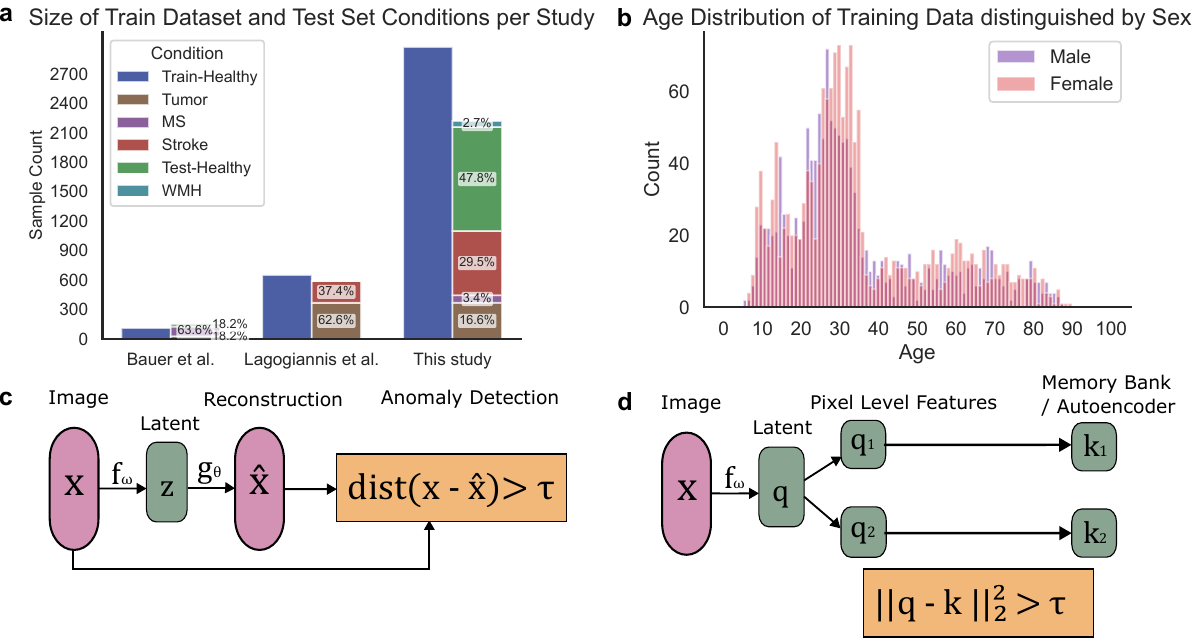}
    \caption{\textbf{Large-scale benchmarking of interdisciplinary state-of-the-art deep unsupervised anomaly detection. }\textbf{a)} Number of volumes in the training and test datasets, split by condition, compared to previous benchmarks. \textbf{b)} Age distribution of the training data. \textbf{c)} Example of a \textit{reconstruction-based} approach. An encoder and decoder are trained exclusively on healthy data, and during testing anomalies are identified by thresholding the reconstruction error with $\bm{\tau}$. \textbf{d)} Example of a \textit{feature-based} approach. A feature map is created by passing the image through a neural network. These features are either (i) reconstructed using an autoencoder, analogous to panel c, or (ii) compared directly to features from a memory bank of healthy images. Anomalies are flagged when the Euclidean distance to healthy features exceeds $\bm{\tau}$.}
    \label{fig:Dataset} \end{figure*}

The human brain is the most complex organ in the body, with billions of neurons forming intricate networks that support cognition, behavior, and perception \cite{kandel2000principles, gazzaniga2009cognitive, dayan2005theoretical}. Neuroimaging provides insights into this complexity but necessarily compresses rich biology into high-dimensional, often noisy measurements. Diseases of neurodegeneration, mental disorders, stroke, and tumors are characterized by substantial heterogeneity at the brain level \cite{davnall2012assessment}. This heterogeneity arises not only from the underlying pathology itself but is further amplified by demographic factors such as age and sex, genetic predispositions, environmental influences, and lifestyle factors. Moreover,  technical variability stemming from magnetic resonance imaging (MRI) acquisition protocols, scanner hardware, and image quality \cite{wachinger2021detect, jovicich2009mri} adds another layer of complexity, making robust characterization and analysis particularly challenging. In clinical practice, diagnosis of brain lesions and treatment planning generally rely on human evaluation of MRI, including lesion identification \cite{thompson2018diagnosis, bakas2017advancing}.  Yet, even for experienced clinicians, the heterogeneity of disease-related changes makes manual assessment time-consuming and error-prone \cite{zijdenbos1994morphometric}. To reduce misdiagnosis, streamline planning, enable longitudinal monitoring, and alleviate workload, automated lesion detection and decision-support systems have emerged \cite{isensee2018nnu,cerri2021contrast}. State-of-the-art systems typically use supervised deep learning to map images to manually annotated lesion masks. While the performance on curated datasets can be strong, generalization to unseen lesion types and different scanners remains limited \cite{xu2024feasibility, ghafoorian2017transfer}, and assembling large expert-annotated datasets is costly \cite{galbusera2024image}. This tension underscores the need for methods that reduce dependence on manually created lesion maps. Such approaches must not only generalize across diverse lesion types, sizes, and contrasts but also adapt to the wide range of individual brain anatomies and disease presentations observed in clinical populations while remaining robust to demographic variability and nuisance variables.

Unsupervised Anomaly Detection (UAD) has emerged as a widely adopted alternative to supervised approaches in MRI analysis and lesion detection. The central principle of UAD is to model a representation of normal brain structure using clean data and to flag deviations from this learned concept of normality as potential anomalies in unseen cases \cite{ruff2021unifying}. By shifting the focus from predefined lesion labels to deviations from normative patterns, UAD provides a framework that can potentially capture unexpected, subtle, or previously uncharacterized pathologies, making it particularly valuable for heterogeneous and poorly understood brain disorders. In brain MRI, recent deep UAD methods typically fall into two main categories: (i) \textit{reconstruction-based} approaches, which learn to compress and reconstruct healthy images and then use voxel-level residuals between the input and reconstruction to localize anomalies \cite{bercea2024diffusion,liang2024itermask,frotscher2023unsupervised} , and (ii) \textit{feature-based} approaches, which extract intermediate representations from neural networks and apply a secondary detection model for pixel-level anomaly localization \cite{roth2022towards,you2022unified,deng2022anomaly}.
Furthermore, training strategies based on synthetic-anomalies have been introduced to further reduce reliance on real lesion maps by altering images of healthy individuals. Often, these approaches incorporate \textit{reconstruction-based} methods to mitigate either detection or generalization problems \cite{zavrtanik2021draem,naval2024ensembled}.

Analogous unsupervised anomaly detection techniques have already been established in other domains, such as industrial surface defect detection \cite{bergmann2019mvtec,roth2022towards,you2022unified,deng2022anomaly}, autonomous driving \cite{hendrycks2019scaling}, and time series analysis \cite{blazquez2021review, audibert2020usad}, where standardized datasets and evaluation protocols have catalyzed progress. In contrast, progress in medical imaging has long been hampered by the lack of comparable large-scale resources. Only with the advent of initiatives such as the Human Connectome Project \cite{van2012human, somerville2018lifespan, bookheimer2019lifespan} and UK Biobank imaging \cite{miller2016multimodal} has this begun to change. Perhaps unsurprisingly, as medical data is inherently sensitive, it is more difficult to acquire, share, and pool at scale \cite{rieke2020future}. To compensate, some studies have resorted to practices utilizing a “healthy” reference from clean slices of anomalous volumes. This strategy potentially introduces systematic biases and undermines the generalizability and robustness of findings, as non-anomalous slices in brains with lesions can hardly be considered to belong to a healthy cohort of brains \cite{bercea2023bias}. Beyond the challenge of reference definition \cite{hajek2007reference}, evaluations in medical imaging often differ in fundamental aspects, including performance metrics, lesion taxonomies, brain regions analyzed, preprocessing strategies, and the degree of control for scanner and demographic factors that render the comparison of methods impossible. Although valuable steps toward unification have been made \cite{baur2021autoencoders, lagogiannis2023unsupervised}, these efforts remain fragmented and leave critical aspects inconsistently addressed, such as evaluation with healthy data, the diversity of the lesions tested, the impact of demographics on predictions, and missing thresholding strategies to derive the predictions. As a result, a stringent and clinically grounded benchmark is needed to establish the current state of the field, create a fair basis for comparison, and guide the development of more robust and generalizable methods in the future.

To address this need, we introduce a comprehensive, multi-center benchmark of deep UAD for brain MRI (see \cref{fig:Dataset}). Our training cohort is five times larger than that in previous studies and includes scans across a large age range and both sexes. Evaluation spans a holdout set of healthy individuals, diverse lesion types, and complex diseases while explicitly assessing the effects of different scanners and demographics (in particular, age and sex). Alongside leading and established UAD approaches developed for medical image analysis, we include industrial surface defect detection methods to test cross-disciplinary insights and enable direct comparisons of state-of-the-art techniques across fields. Our evaluation of stroke, tumors, multiple sclerosis (MS), and white matter hyperintensities (WMH) demonstrates that \textit{reconstruction-based} approaches, many of which are based on diffusion models, achieved the strongest overall performance, while \textit{feature-based} methods showed greater robustness to distributional shifts. At the same time, systematic biases linked to lesion size, image contrast, age, sex, and scanners remain major obstacles to clinical translation.

\section{Results}
\label{sec:Results}

\begin{figure*}[ht]
    \centering
    \includegraphics[scale=0.75]{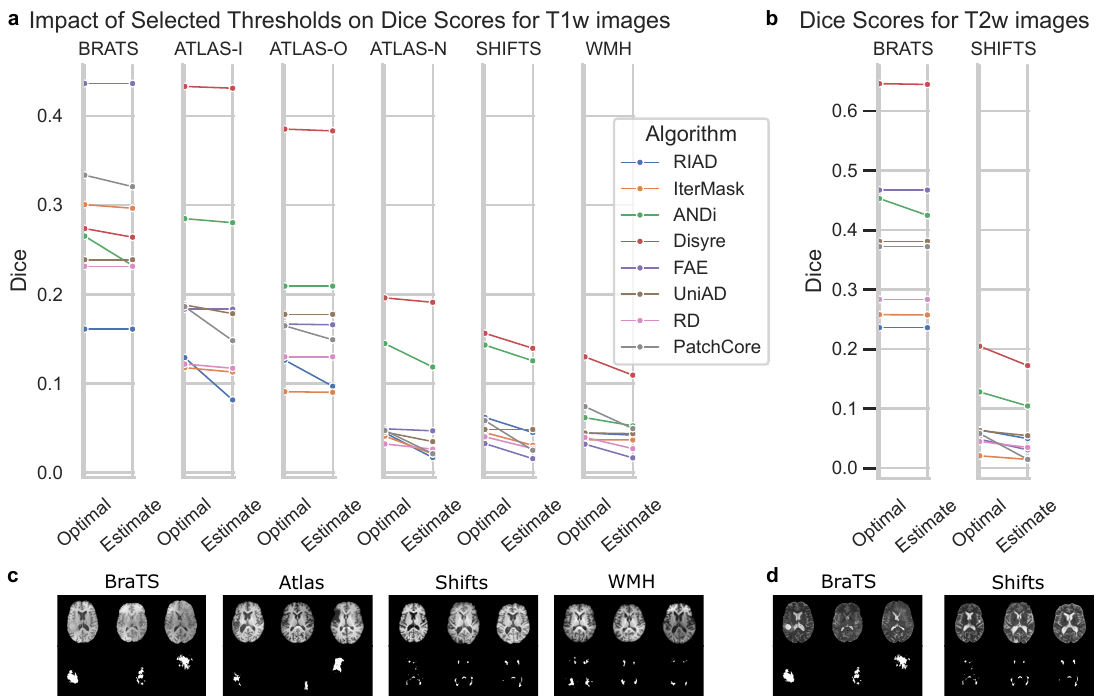}
    \caption{\textbf{Multi-site and multi-task evaluation of state-of-the-art unsupervised anomaly detection a)} The performance of the algorithms on T1w images were reported using two different thresholds, the \textit{optimal} threshold, defined as the maximum possible Dice score optimized in the test set, thus potentially susceptible to bias but standard in the field. Second, the \textit{estimated} threshold optimized on the validation data set, then fixed and performance for that threshold reported on the untouched test set, thus unbiased but not the standard in the field. For the large lesions we did not observe a substantial difference between the thresholding procedures.  \textbf{b)} The performance of the algorithms with the \textit{optimal} and \textit{estimated} thresholds on structural T2 weighted images. \textbf{c-d)} Example images and labels for each dataset and modality. Taken together, we show heterogeneous performance across lesions and modalities, no single method was superior across tasks. Across all evaluations the best performing methods were Disyre followed by ANDi.} \label{fig:Evaluation}
\end{figure*}

We trained six different models belonging to the broader category of \textit{reconstruction} and \textit{feature-based} methods.  All models described in \cref{sec:Models} were trained separately using T1-weighted (T1w) and T2-weighted (T2w) MRIs. We used 2976 T1w scans and 2972 T2w from healthy individuals, which amounts to approximately 461,000 imaging slices aggregated from multiple datasets for training. Details regarding the data are described in \cref{Table1} and \cref{sec:Data,sec:Pre}. After training, the models represent a normative concept that can be used to calculate deviations at the voxel level and are referred to as anomaly maps. 

The calculated anomaly maps are continuous, and a threshold $\bm{\tau}$ is needed for the binary classification of a voxel as either healthy or anomalous. In the literature, this threshold is usually not estimated \cite{bercea2024diffusion,liang2024itermask,meissen2022unsupervised,frotscher2023unsupervised}, and the test set, which typically contains only a single lesion type, is used to determine the \textit{optimal} $\bm{\tau}$, potentially leading to inflated benchmarks \cite{zimmerer2019unsupervised}. In a clinical scenario, this corresponds to instances in which the condition is already known. By contrast, we estimated the threshold by using a separate validation dataset that contains all lesion types observed during the test phase, including tumors, chronic stroke, multiple sclerosis (MS), and white matter hyperintensities (WMH). Specifically, the threshold is tuned for each method by finding the best mean Dice score, which results in an unbiased threshold based on optimal performance in the validation set concerning lesion detection. Note that tuning the hyperparameters or the threshold using anomaly data is in contrast to unsupervised learning, but is the standard approach for deep learning methods in anomaly detection \cite{vskvara2018generative}.

We evaluated all implemented algorithms with the \textit{optimal} and \textit{estimated} thresholds on a highly heterogeneous collection of lesions spanning multiple pathologies, sizes, contrasts, and distributions, and we refer the reader to \cref{subsec:Dice} for the results. In addition to the evaluation of volumes containing lesions, the performance of the algorithms at the \textit{estimated} threshold was evaluated on hold-out healthy data in \cref{subsec:FPR}. In \cref{subsec:Scanner}, the effects of domain shifts, lesion size, and their interaction on detection performance were examined. In \cref{subsec:Age}, the effects
of demographic variations were systematically evaluated. Lastly, the influence of data abundance for reference definition has been examined in \cref{subsec:Scale}.   
The concept of the benchmark and the categories of the methods used are shown in \cref{fig:Dataset}. For more details regarding the individual methods and the data used, we refer the reader to \cref{sec:Methods}. 

\subsection{Algorithmic Performance Across Brain Lesions}
\label{subsec:Dice}
Among all evaluated methods, Disyre\cite{naval2024ensembled} achieved the highest overall performance, followed by ANDi\cite{frotscher2023unsupervised} and FAE\cite{meissen2022unsupervised} (see \cref{fig:Evaluation}). In general, algorithms performed better on larger lesions, on hyperintense lesions (e.g., in T2w images), and on in-distribution (ID) samples. By contrast, \textit{feature-based} methods consistently struggled to detect small lesions such as MS and WMH, as well as the predominantly small lesions in the ATLAS-N dataset. Interestingly, the same methods appeared more robust to out-of-distribution (OOD) samples in ATLAS-O, suggesting a trade-off between sensitivity to small, subtle lesions and robustness to distributional shifts. Threshold selection emerged as a critical factor for performance, particularly for smaller lesions (MS, WMH, ATLAS-N), indicating that deviations are shaped not only by lesion type but also by lesion size. For larger lesions, threshold sensitivity was more limited and was primarily observed in a subset of algorithms, such as PatchCore, ANDi, and RIAD. This finding underscores the importance of realistic evaluations and suggests that adaptive thresholding strategies based on the specific individual or their demographics could improve future methods.

\subsection{Algorithmic Performance Across Healthy Variations}
\label{subsec:FPR}
\begin{figure}[ht]
    \centering
    \includegraphics[scale=0.27]{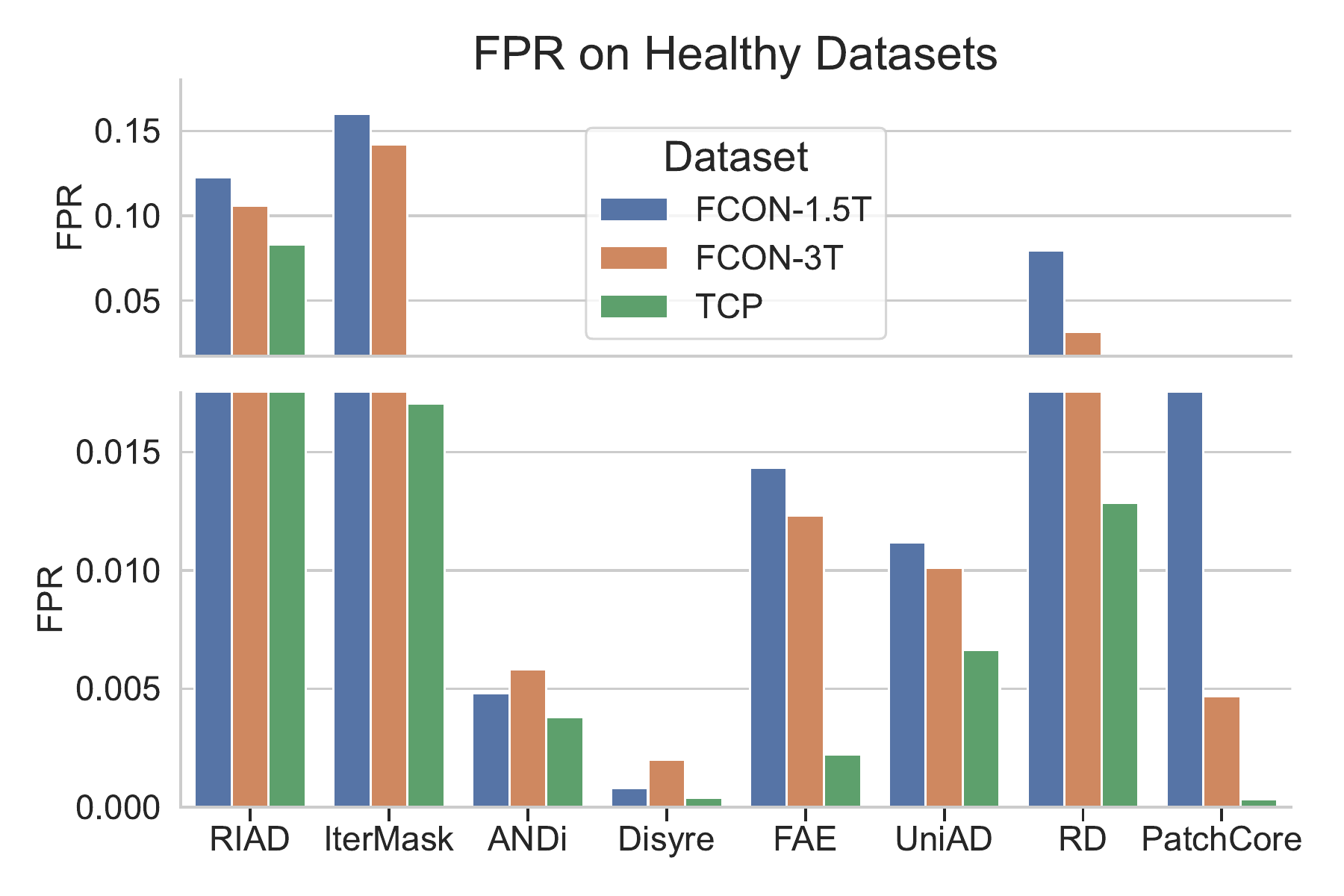}
    \caption{\textbf{False positive rate on healthy brains across all tested algorithms associated with \textit{estimated} threshold.} For some methods we identified high false positive rates and heterogeneous performances across the healthy cohort for T1w images (for T2w see supplement). This shows that the methods were biased for either imaging protocols or demographics and selecting the right decision threshold is critical.}
    \label{fig:Healthy}
\end{figure}

In classical anomaly detection, the maximum acceptable false positive rate is predefined, and the threshold is chosen accordingly\cite{liang2017enhancing,goyal2020drocc}. Here, we optimized the threshold using the Dice score on the validation dataset, as it balances precision and recall and is the standard metric for medical image segmentation \cite{muller2022towards}. Since different lesion types (tumors, stroke, MS, WMH) each have distinct ideal balances with the negative class (healthy voxels), we used a validation dataset containing all lesion types to define an unbiased operating point with meaningful detection accuracy and low false positive rates. We evaluated the resulting false positive rate in the healthy cohort to understand how the realistic operating point influences false alarms for healthy individuals under domain shifts that could not have been accounted for with the ideal balance in the validation dataset. Three datasets containing only healthy individuals, imaged with different protocols, were used to evaluate the false positive rate of the algorithms. Note that these datasets represent the general population, where lesions are extremely rare, and samples are generally lesion-free. Applying our algorithms to such data, therefore, provides a good estimate. As shown in \cref{fig:Healthy}, all algorithms performed differently across the three distinct datasets, indicating a strong influence from either imaging protocols or demographics. Overall, the methods showed the highest false positive rates (FPRs) on the 1.5T FCON scans and the best performance on TCP. We attribute the reduced performance on FCON-1.5T to lower image quality and the limited number of 1.5T scans in the training data, while FCON-3T likely includes OOD samples from unseen scanners or protocols. TCP scans belong to the ID distribution, which could explain the higher performance. In general, FPRs in healthy samples correlated with lesion detection performance. Disyre performed best, followed by ANDi and PatchCore. Interestingly, some \textit{feature-based} approaches (e.g., PatchCore, UniAD) achieved lower FPRs but exhibited weaker detection, and their performance varied significantly across datasets, indicating bias from other factors. For T2w scans, FPR values were computed on TCP (Suppl. Fig. S1). Importantly, this analysis highlights threshold selection as a key bottleneck for clinical translation: despite using a diverse validation dataset, the thresholds produced large FPR differences across healthy datasets.

\subsection{Impact of Domain Shifts and Lesion Load Variability}
\label{subsec:Scanner}

\begin{figure*}[ht]
    \centering
    \includegraphics[scale=0.8]{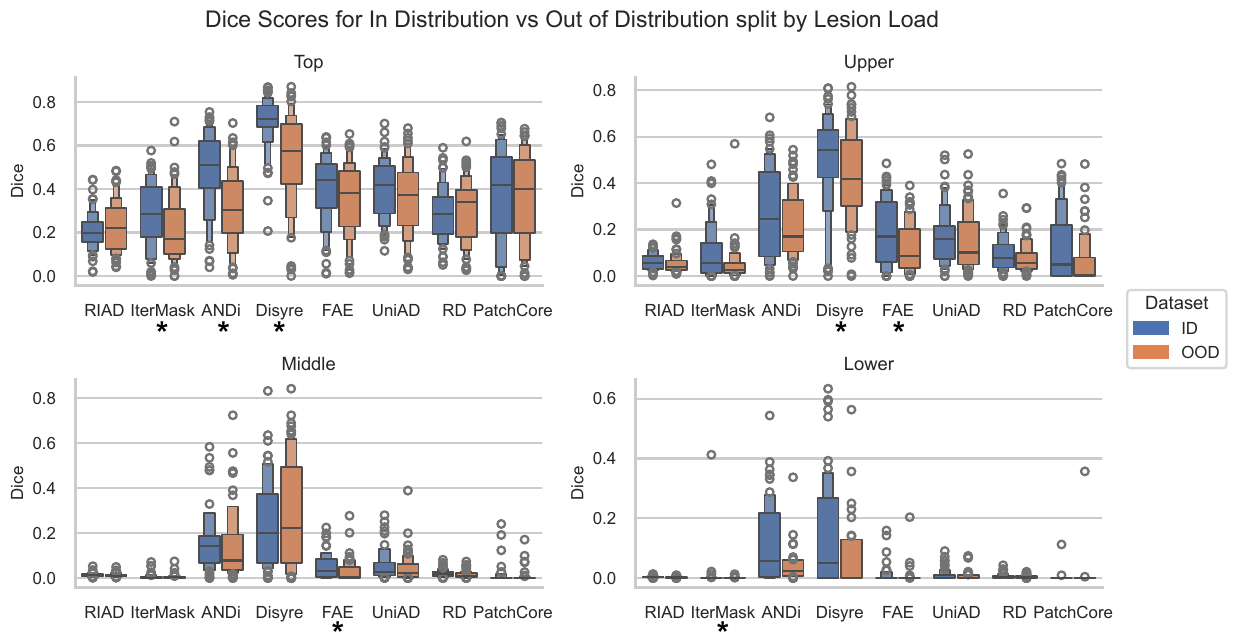}
    \caption{\textbf{Out-of-distribution effects due to scanner differences and influence of lesion load.} The ATLAS dataset had been split by the percentiles of the lesion load. Top corresponds to a lesion load above the 75th percentile, upper - above 50th percentile to 75th, middle - above 25th percentile to 50th percentile, lower - below 25th percentile. Columns marked with \textbf{*} correspond to the distributions of Dices scores that are different according to the Mann-Whitney U test after multiple testing correction with the Benjamini–Hochberg adjusted significance level $\alpha=0.05$. Taken together,  \textit{reconstruction-based} approaches are more sensitive to imaging protocols, while \textit{feature-based} methods are generally robust to this source of variation in the data.}
    \label{fig:Scanner}
\end{figure*}

Robustness to domain shifts, particularly those induced by differences in MRI scanner hardware, is essential for clinical translation. Using the ATLAS dataset, we partitioned the data into ID and OOD subsets, stratifying cases by lesion load, i.e., number of anomaly voxels in the volume, to investigate how lesion burden interacts with domain variability (see \cref{fig:Evaluation,subsec:Dice}). We constructed eight datasets corresponding to four matched ID–OOD pairs, stratified by lesion load percentiles derived from the full ATLAS cohort. We group volumes above the 75th percentile, those above the 50th percentile to the 75th, those above the 25th percentile to the 50th percentile, and those below the 25th percentile. This design ensured that each pair contained scans with comparable lesion burdens, thereby isolating the effects of scanner heterogeneity from lesion size. Statistical evaluation using Mann–Whitney U tests with Benjamini–Hochberg correction revealed a significant decline in segmentation performance for most \textit{reconstruction-based} methods when applied to OOD data with high lesion burden, while \textit{feature-based} methods remained largely unaffected, as shown in \cref{fig:Scanner}. The only exception was FAE, which showed significant performance losses in the upper and middle lesion load percentiles. Across all methods, Dice scores decreased with decreasing lesion load, where lesions are smaller and more subtle, thereby challenging algorithmic sensitivity, and the performance gap between ID and OOD samples narrowed for smaller lesions, indicating that lesion size exerts a stronger influence on model performance than scanner-related factors. Nevertheless, even top-performing algorithms displayed marked drops in the top and upper lesion load strata under domain-shift conditions, exposing critical vulnerabilities to scanner differences and lesion variability. The p-values and corrected p-values can be found in Suppl. Table 6. To further test the interaction between lesion load and domain variability, we conducted a Scheirer-Ray-Hare test with the lesion load categories and the ID-OOD dummy variable as the independent variables to predict the Dice scores. All algorithms were significantly impacted by lesion load; ANDi and Disyre were impacted by the ID-OOD split, and Disyre exhibited a significant interaction term. The degrees of freedom and p-values for the Scheirer-Ray-Hare tests can be found in Suppl. Table 7. These results underscore the urgent need for models that generalize reliably across acquisition settings and patient populations, particularly in scenarios with small, clinically relevant lesions.

 \subsection{Impact of Age and Sex on false positive rates and Lesion Identification}
 \label{subsec:Age}

 \begin{figure*}[ht]
    \centering
    \includegraphics[scale=0.73]{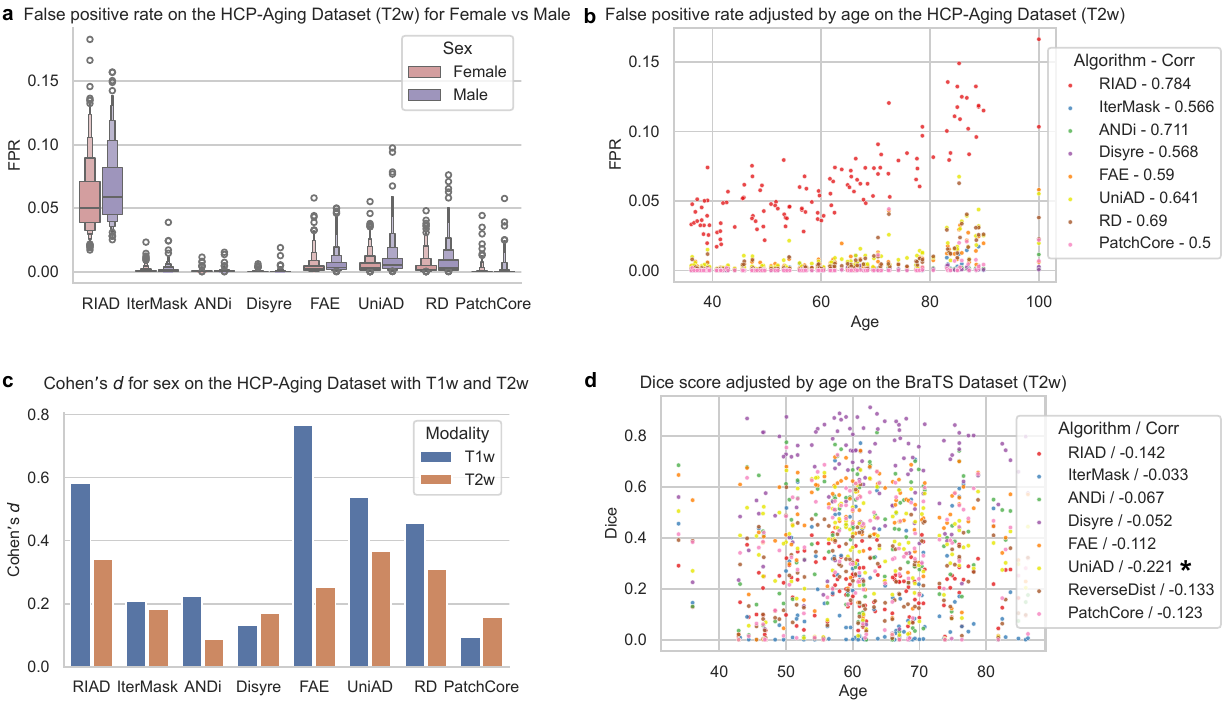}
    \caption{\textbf{Impact of demographics on false positive rates and lesion identification. a)} The false positive rate on the HCP Aging dataset for the female and male group. All groups show significant differences based on the Mann-Whitney U test. Significance level $\alpha=0.05$ was used for all tests.  \textbf{b)} The false positive rate on the HCP Aging dataset adjusted by age. The Spearman rank correlation test displayed that all algorithms show significant positive correlation indicating that older individuals are more likely to have an anomaly assigned to a voxel that is normal. \textbf{c)}  Cohen’s~\textit{d} on the HCP Aging dataset for the groups male and female with both modalities. All algorithms show a positive Cohen’s~\textit{d}, an effect due to higher false positive rates for males. \textbf{d)} The Dice score on the BraTS dataset adjusted by age. Only UniAD (marked with \textbf{*}) shows a significant negative correlation (Spearman rank correlation). Bias for age and sex on healthy volumes generally decreased with increasing performance of the algorithm.}
    \label{fig:Age}
\end{figure*}
 
The performance of unsupervised anomaly detection methods varied substantially across individuals due to multiple interacting factors. Lesion size was a key determinant of detection accuracy, with larger lesions producing stronger deviations and being more readily detected, while false positive rates showed considerable variation across methods and datasets (see \cref{fig:Healthy}). To further analyze this behavior, we studied the impact of inter-individual anatomical variability linked to demographic attributes such as age and sex in the HCP Aging and BraTS datasets. As shown in \cref{fig:Age}, the Spearman rank correlation test with $\alpha=0.05$ indicated that models tended to overestimate the occurrence of anomalies in older individuals in the HCP Aging dataset, likely due to age-related changes in brain structure. Interestingly, there was no effect of age on the Dice score. To evaluate differences between the sexes, we calculated a Mann-Whitney U test on the FPR as a proxy for the fairness of our algorithms. All algorithms showed significant differences in performance for the different sex groups, with male brains being predicted as more anomalous than female brains. Most algorithms exhibited higher effect sizes on the T1w images, as measured by Cohen’s~\textit{d}, indicating stronger sensitivity to sex-related differences in this modality. Importantly, the top-performing algorithms were among those with the lowest bias, suggesting that high overall accuracy can imply, in our scenario, better fairness across biological groups. Detailed results, including the means and standard deviations of the FPR values for both groups, the corresponding Cohen’s~\textit{d} values, and the full plots for the T1w analyzes, are provided in Suppl. Fig. S2. These findings highlight the need to develop and rigorously validate anomaly detection models that are robust against biological variability. Such resilience is critical not only for the reliable detection of lesions but also for ensuring stable performance in healthy cohorts, thereby supporting both clinical applicability and generalizability across diverse populations.

 \subsection{Effect of Data Scaling During Training}
 \label{subsec:Scale}
\begin{figure*}[ht]
    \centering
    \includegraphics[scale=0.8]{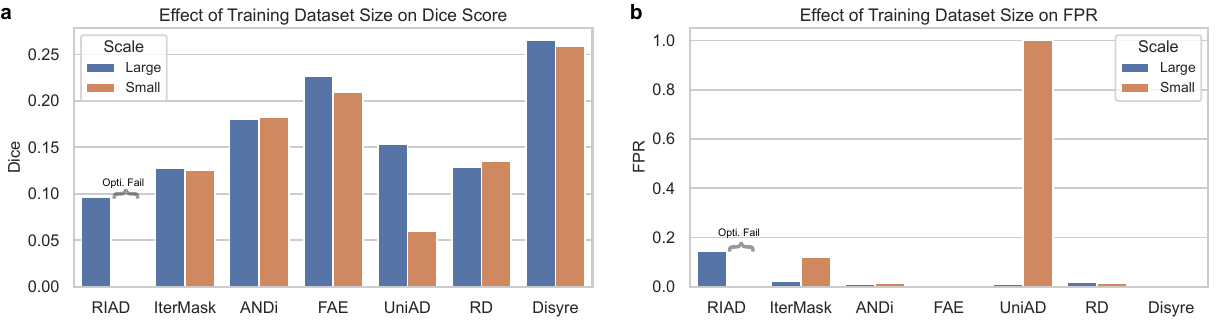}
    \caption{\textbf{Impact of dataset scaling on algorithmic performance a)} The Dice score on the T1w validation dataset. \textbf{b)} The false positive rate for the T1w images in the TCP dataset. The training scale is defined as the number of individuals in the dataset. The small training scale uses 3\% (92 volumes) of the training dataset. This analysis showed that the majority of algorithms are irresponsive to more data, highlighting the need for methods that are capable to grain performance when integrating large scale data for anomaly detection.}
    \label{fig:Scale}
\end{figure*}

To assess the influence of data abundance on performance, we tested the impact of diminished training data and experimented with an additional training dataset that comprised 3\% of the described training dataset used in the main analysis. For the performance evaluation of the newly trained models in the detection task, only the validation dataset was used due to computational constraints. As shown in \cref{fig:Scale}, all algorithms except UniAD performed almost identically in terms of the mean Dice score on the validation dataset. Similarly, performance on false positive rates measured with the TCP dataset was stable across training datasets, with two exceptions: UniAD and IterMask. These results suggest that straightforward scaling strategies alone, by enlarging training data, are insufficient to overcome fundamental limitations for most unsupervised anomaly detection tasks. The analysis was performed on the T1w images, and RIAD was excluded due to difficulties encountered during the optimization procedure for all conducted experiments.

\section{Discussion}

This study presents one of the most comprehensive evaluations of UAD for brain MRI to date, spanning multiple scanners, demographic groups, and four major brain diseases, ranging from stroke and tumors to multiple sclerosis (MS) and white matter hyperintensities (WMH). While recent advances demonstrate clear progress, current approaches remain far from achieving the robustness, sensitivity, and fairness required for clinical deployment \cite{kelly2019key}. \textit{Reconstruction-based} methods, particularly diffusion-inspired approaches such as Disyre and ANDi \cite{naval2024ensembled,frotscher2023unsupervised}, achieved the highest segmentation accuracy, especially for large and hyperintense lesions. \textit{Feature-based} methods, by contrast, were more resilient to scanner variability but consistently struggled with subtle or small anomalies. Several newly developed \textit{reconstruction-based} methods have achieved state-of-the-art performance, but no algorithm achieves clinically relevant performance defined by sufficient detection performance, robustness to domain shifts caused by imaging protocols and potentially new populations, and without unintended discriminatory biases \cite{kelly2019key}. On the contrary, our findings highlight systematic sources of bias, namely that lesion size and image contrast were dominant drivers of accuracy, and false positive rates in healthy cohorts were systematically influenced by demographics, particularly age and sex. Taken together, these results emphasize that future advances must go beyond incremental performance gains. Methodological priorities include increasing sensitivity to small and low-contrast lesions, improving robustness to distributional shifts across scanners and protocols, reducing demographic biases, and refining evaluation procedures to more closely mirror real-world conditions.

A central theme that emerges from this study is the critical role of thresholding and evaluation practices. Because UAD produces continuous anomaly maps, binarization requires a decision threshold. While threshold-independent evaluation exists, it does not provide an indication of the performance under a specific binary decision that is needed for any future diagnosis or longitudinal monitoring; therefore, it is less valuable in a medical context. In much of the literature, thresholds are optimized directly on the test set, potentially resulting in inflated performance estimates that would not hold in clinical settings. In contrast, we determined thresholds using a heterogeneous validation set encompassing multiple lesion types and then evaluated them on unseen test data. This more principled procedure yielded clinically realistic performance estimates that showed differences from standard thresholding procedures for some of the compared models (see \cref{fig:Evaluation}). Such effects underscore that thresholds are not simple post-hoc technicalities but integral components of the detection pipeline. Effective evaluations and eventual deployment, therefore, require adaptive thresholding strategies potentially conditioned on scanner, protocol, or demographic information, as well as standardized validation procedures to ensure fair and reproducible comparisons across methods. 

Closely related to this issue are the implicit assumptions underpinning \textit{reconstruction-based} approaches, which we deem to be a major cause of their shortcomings with respect to clinical usability. The first assumption is that a model trained exclusively on healthy data will always reconstruct a healthy sample. While diffusion-based methods partly relax this assumption through the forward process, enabling a tradeoff between reconstruction fidelity and an image from the data distribution, interference in this process is practically non-trivial \cite{bercea2023mask, frotscher2023unsupervised}. Furthermore, control of the reverse process to create conditional samples remains mathematically opaque for diffusion and flow models \cite{lipman2024flow}.  More classical methods, such as RIAD or IterMask, employ complex masking strategies to encourage useful reconstructions; however, these do not guarantee that anomalies are removed. Even subtle intensity perturbations can suffice when residual error is used. The second assumption is that the deviation measure between input and reconstruction is inherently suitable for anomaly localization. The commonly used residual error has well-documented drawbacks \cite{meissen2021pitfalls}, as it treats each pixel independently and ignores spatial context. Alternatives such as the Structural Similarity Index Measure (SSIM), as used in the FAE \cite{meissen2022unsupervised}, partly address this by incorporating locality. However, no deviation metric or measure has yet been explicitly designed for brain MRI. Therefore, we suggest that the choice of deviation is as critical as the reconstruction itself, and developing neuroanatomically informed measures could represent a key avenue for advancing \textit{reconstruction-based} UAD.

In contrast to previous reports \cite{lagogiannis2023unsupervised}, \textit{feature-based} approaches did not achieve the highest performance in our benchmark. Consistent with earlier analyzes, however, we observed that these methods were particularly weak at detecting small lesions. We hypothesize that this relative decline reflects the rapid advances in \textit{reconstruction-based} and generative modeling approaches, which have shifted the comparative performance landscape \cite{zavrtanik2021reconstruction, liang2024itermask,frotscher2023unsupervised,naval2024ensembled,meissen2022unsupervised,you2022unified,deng2022anomaly,roth2022towards}. The central assumption underlying \textit{feature-based} frameworks is that pretrained features provide sufficient representational power to distinguish anomalies from normal tissue in the embedding space. All \textit{feature-based} methods included here relied on ImageNet-pretrained networks, even when originally developed for brain MRI (e.g., FAE). While such features can yield competitive performance, the use of ImageNet pretraining is suboptimal, and the extension to MRI-specific pretraining appears to be a natural progression. Indeed, prior work suggests that MRI-based pretraining can yield small but consistent improvements \cite{lagogiannis2023unsupervised}. We did not pursue this option in the present benchmark, as no recent UAD method has demonstrated state-of-the-art performance using MRI-pretrained features. The scarcity of such approaches in the literature likely reflects both practical and conceptual barriers. One important barrier is the lack of MRI datasets that are comparably large and harmonized to the scale of ImageNet \cite{ahmed2023systematic}. Another difficulty is the design of pretext tasks that transfer effectively to the downstream challenge of anomaly detection. Overcoming the former barrier is likely infeasible, but the latter could reinvigorate \textit{feature-based} pipelines and potentially restore their competitiveness, especially if paired with strategies that address scanner heterogeneity and small-lesion sensitivity.

Sensitivity to domain shifts caused by scanner effects emerged as an important take-home message as a result of our benchmark. By stratifying the ATLAS dataset into in-distribution and out-of-distribution subsets matched for lesion load, we were able to disentangle the effects of scanner variability from lesion size. \textit{Reconstruction-based} methods showed pronounced degradation on OOD data, particularly when lesion burdens were high, suggesting that larger lesions may amplify reliance on scanner-specific statistical cues. \textit{Feature-based} methods were generally more stable across scanners. However, exceptions such as the FAE model highlight how deviation measures strongly influence generalization. We suspect that the FAE’s susceptibility to OOD shifts arises from its use of an image-derived deviation measure, which treats high-resolution feature maps as if they were natural images and therefore inherits scanner sensitivities. Importantly, our results reveal that lesion size and domain shifts can interact, with large lesions amplifying vulnerability to acquisition-related biases, whereas small lesions are inherently more difficult to detect. These findings emphasize that robust generalization to heterogeneous imaging protocols remains an unsolved challenge for any possible future deployment. Addressing this issue will require explicit domain adaptation strategies, including scanner-aware harmonization \cite{wachinger2021detect}, training on more diverse acquisition protocols, and potentially test-time adaptation to local distributions to ensure consistent performance across imaging environments. Demographic factors also had a systematic effect. False positives increased with age in healthy cohorts, reflecting normal age-related changes misclassified as pathology. All algorithms showed significant sex-related differences, with male brains more often flagged as containing anomalous regions. Although the top-performing methods exhibited somewhat reduced demographic bias, these results underscore the need for fairness-aware modeling \cite{bercea2023bias}. Incorporating demographic factors into normative reference frameworks \cite{rutherford2022normative} and anomaly detection workflows, explicitly quantifying subgroup biases \cite{rutherford2024reference}, and potentially calibrating thresholds in a stratified manner will be essential to ensure equitable performance. More broadly, these observations point to the necessity of developing models that are resilient to natural anatomical diversity and capable of distinguishing genuine pathology from normative variability.

We observed a limited impact of scaling training data. Whereas supervised machine learning typically benefits substantially from larger datasets \cite{hestness2017deep,desai2019technical}, simply adding more healthy scans yields only minor improvements in UAD performance for some algorithms. This reflects the algorithmic constraints of the UAD approaches, for which training solely on healthy cohorts cannot effectively teach models which deviations are clinically meaningful. Additionally, subtle lesions are often indistinguishable from normal anatomical variation. In line with prior work \cite{lagogiannis2023unsupervised}, our experiments confirmed that even when the training data were drastically reduced, most methods showed comparable lesion detection accuracy and false positive rates for in-distribution evaluations. These results suggest that naive scaling is insufficient to overcome the intrinsic challenges of neuroimaging anomaly detection. Future progress will require qualitatively new strategies, including anatomically informed deviation measures, large-scale medical pretraining with task-aligned self-supervised objectives, and robust domain adaptation frameworks that explicitly account for clinical variability.

Although our study drew on a broad range of datasets spanning multiple lesion types and imaging protocols, the collection cannot fully reflect the diversity of real-world clinical neuroimaging. Rare pathologies \cite{lee2020brain}, pediatric cohorts \cite{somerville2018lifespan}, and cases acquired under extreme or atypical imaging conditions remain underrepresented, which may limit the generalization of our conclusions. Similarly, although we explicitly examined the demographic effects of age and sex, other important axes of biological and social variability were not captured. Factors such as ethnicity, comorbidity, medication status, or socioeconomic background can shape brain anatomy and imaging characteristics \cite{llera2019inter}, and their influence on anomaly detection remains unexplored. Another limitation lies in our choice of evaluation targets. Thresholds were optimized with respect to segmentation accuracy, which provides a clear and widely accepted benchmark but may not align with clinical decision-making needs. Tasks such as patient-level triage, prognosis, or monitoring treatment response may require different operating points or evaluation metrics, emphasizing sensitivity, specificity, or predictive value over voxel-level overlap. Future studies should, therefore, consider multiple evaluation frameworks that reflect the range of clinical contexts in which anomaly detection could be applied. Methodological scope is another constraint. Although we implemented a broad set of state-of-the-art algorithms, not all contemporary or emerging approaches could be included. Having said that, we implemented relevant and current state-of-the-art algorithms while aiming for algorithmic diversity in the methods used. Moreover, although our pipeline was carefully designed to minimize data leakage and benchmark bias \cite{varoquaux2022machine}, it still cannot reproduce all conditions relevant to clinical deployment. In practice, anomalies are rare, heterogeneous, and often subtle, and scans are embedded in complex diagnostic workflows. These realities pose additional challenges, such as handling incidental findings, balancing false positives against clinical workload, and integrating uncertainty estimates that remain outside the scope of this study. Taken together, these limitations highlight that while our benchmark offers an important step toward the systematic evaluation of UAD in neuroimaging, further work is needed to expand dataset diversity, incorporate broader demographic and clinical factors, refine evaluation criteria, and explore algorithmic strategies in conditions that more closely approximate real-world deployment.

Looking forward, several priorities emerge for future research. First, the design of principled deviation metrics should become a central focus. Residual errors and generic SSIM-based measures have well-documented limitations \cite{meissen2021pitfalls, wang2003multiscale}, and future work should aim for neuroanatomically grounded deviations. Having said that, this task is admittedly extremely difficult, and determining whether this ambition is feasible is still subject to current research \cite{koval2017statistical, delmonte2019white, krebs2019learning, balakrishnan2019voxelmorph}. Benchmarks like ours, along with further developments using the principles outlined here, will allow us to evaluate such methods in more depth than is currently common.  Second, large-scale medical pretraining on curated neuroimaging datasets, combined with meaningful self-supervised tasks, could provide domain-native representations that are more sensitive to subtle anomalies than ImageNet features. Third, harmonization and domain adaptation need to be built directly into the modeling pipeline to ensure robust performance across scanners and acquisition protocols. Fourth, fairness-aware modeling should be prioritized \cite{bercea2023bias}, with systematic evaluation of demographic biases and strategies to mitigate them. Finally, the evaluation itself must incorporate thresholds, as outlined here, and the reporting of uncertainty, robustness, and fairness alongside accuracy is essential to establish clinical trust.

In conclusion, this benchmark underscores both the promise and the current limitations of UAD in brain MRI. While modern approaches can detect a broad range of lesions, their performance is uneven across lesion types, scanners, and populations. They remain susceptible to bias and thresholding procedures. Addressing these challenges will require innovations that go beyond scaling data or marginal architectural tweaks. By developing principled deviation metrics, MRI-native pretraining, robust domain adaptation, fairness-aware pipelines, and clinically meaningful evaluation frameworks, the field can move closer to reliable, equitable, and actionable anomaly detection in neuroimaging.
 
\section{Methods}
\label{sec:Methods}
Here, we provide information on the data (see \cref{sec:Data} and \cref{Table1}), preprocessing (\cref{sec:Pre}), and the evaluated methods (see \cref{sec:Models} and \cref{Table2}). Additional information about training and tuning for all methods can be found in the supplementary methods.

\subsection{Datasets}
\label{sec:Data}
\begin{table*}[t]
    \centering
    \caption{MRI Details of Multi-Side Data}
    \vspace{0.5cm}
    \begin{tabular}{ |p{2cm}||p{2cm}|p{2cm}|p{2cm}|p{2cm}|}
    \hline
     \textbf{Dataset}& \textbf{\# Scans} &\textbf{\# Scanners}&\textbf{Modalities}&\textbf{Condition}\\
     \hline
     CAMCAN   &652    &1&T1w, T2w& Healthy\\
     HCP 1200&1088  &1&T1w, T2w&Healthy\\
     HCP Dev &651 &1&T1w, T2w& Healthy\\
     IXI    &580&3&T1w, T2w&Healthy\\
     \hline
     TCP&92&1&T1w, T2w&Healthy\\
     FCON&244& Unknown&T1w& Healthy\\
     HCP Aging&725&1&T1w, T2w&Healthy\\
     BraTS&369&Unknown&T1w, T2w& Tumor\\
     Atlas&655&14&T1w&Stroke\\
     Shifts&76&Unknown&T1w, T2w&MS\\
     WMH&60&3&T1w&WMH\\
     \hline
    \end{tabular}
    \label{Table1}
\end{table*}

For training, we used large-scale healthy cohorts, including the Cambridge Center for Aging and Neuroscience dataset (CAMCAN) \cite{taylor2017cambridge}, the Human Connectome Project (HCP) Young Adult (S1200) dataset \cite{van2012human}, the HCP Development dataset \cite{somerville2018lifespan}, and the IXI dataset \cite{IXIdataset}, resulting in 2,976 T1w and 2,972 T2w scans acquired across six scanners. For validation, we randomly selected approximately 10\% of the individuals from all other lesion datasets, described below, yielding about 92 T1w and T2w scans. This dataset was used to tune hyperparameters and estimate unbiased thresholds. For testing, we included both healthy and clinical cohorts. The healthy cohort comprised the Transdiagnostic Connectome Project (TCP) \cite{chopra2024transdiagnostic}, a subset of the 1000 Functional Connectomes Project (FCON) \cite{FCONdataset}, and the HCP Aging dataset \cite{bookheimer2019lifespan}. The clinical cohort was made up of the Multimodal Brain Tumor Segmentation Challenge (BraTS 2020) \cite{menze2014multimodal,bakas2017advancing,bakas2018identifying}, the Anatomical Tracings of Lesions After Stroke (ATLAS v2.0) \cite{liew2022large}, the MSSEG and Ljubljana MS lesion datasets from the 2021 Shifts Challenge \cite{shifts2021}, and the White Matter Hyperintensities (WMH) Challenge dataset from MICCAI 2017 \cite{AECRSD_2022}. In total, the test set consisted of 2,221 T1w and 1,262 T2w scans, with the distribution of lesions and healthy samples summarized in \cref{Table1} and visualized in \cref{fig:Dataset}. 
To analyze scanner-related domain shifts, the FCON dataset was divided into 139 scans acquired on 1.5T scanners (München, Oulu, Orangeburg) and 105 scans acquired on 3T scanners (Atlanta, Palo Alto, Ann Arbor). The ATLAS dataset was further partitioned into AtlasI (262 scans from scanners also used during training), AtlasO (227 scans from unseen scanners), and AtlasN (166 scans from scanners with unidentifiable information, which by chance contained mostly smaller lesions). Additionally, the full ATLAS cohort was stratified into four groups (Top, Upper, Middle, and Lower) based on lesion load, using the 25th, 50th, and 75th percentiles, and combined with the scanner-based partitions, resulting in 57 individuals for ID Lower, 34 for OOD Lower, 51 for ID Middle, 58 for OOD Middle, 77 for ID Upper, 58 for OOD Upper, 79 for ID Top, and 77 for OOD Top. These splits allowed us to systematically evaluate algorithm robustness under distributional shifts induced by both scanner variability and lesion burden.

\subsection{Preprocessing}
\label{sec:Pre}
For preprocessing the CAMCAN, IXI, FCON, TCP, ATLAS, and WMH datasets, we applied a pipeline similar to the UK Biobank protocol \cite{alfaro2018image}. Using FSL \cite{jenkinson2012fsl}, all images were reoriented to match MNI152 space \cite{grabner2006symmetric} (rotation only, no registration). For multimodal datasets, rigid transformations to the individual’s T1w image were computed with FLIRT \cite{jenkinson2002improved} and applied to align the modalities. This ensured that all subsequent transformations could be calculated on the T1w and transferred to other contrasts. Because Siemens gradient nonlinearity correction requires proprietary files and not all scans were Siemens acquisitions, this step was omitted. Instead, preprocessing proceeded with a field-of-view reduction on the T1w images, as in the UKB pipeline. Non-linear registration to MNI152 space was performed with FNIRT \cite{andersson2008fnirt} to generate a standard-space brain mask, which was then inverted and applied for skull stripping across modalities. After skull stripping, all scans were rigidly registered to the SRI24 atlas \cite{rohlfing2010sri24} and resampled to 1 mm isotropic resolution. In cases where large anomalies (e.g., ATLAS lesions) rendered FNIRT unreliable, ROBEX \cite{iglesias2011robust} was used for skull stripping. For HCP datasets, we used the minimally preprocessed scans \cite{glasser2013minimal}, additionally applying rigid registration and interpolation to SRI24 at 1 mm isotropic resolution for consistency with other datasets. Shifts challenge data \cite{shifts2021} was already denoised, skull stripped, bias-field corrected, and interpolated; we therefore only applied rigid registration to SRI24. One Ljubljana case with a failed skull stripping was excluded. BraTS scans had undergone skull stripping, rigid registration to SRI24, and interpolation, but with differing orientations; these were reoriented with FSL. It is important to note that the dataset collection included both bias-field–corrected and uncorrected scans, as well as varying skull-stripping procedures across training and test sets. For deep learning preprocessing, all volumes were min–max normalized to [0,1], converted into axial slices, and zero-only slices were removed. The remaining slices were center-cropped to $224\times224$. Algorithm-specific preprocessing exceptions are described in the Supplementary Methods.

\subsection{Models, Architectures and Training}
\label{sec:Models}

\begin{table*}[t]
    \centering
    \caption{Origin and Framework of the Methods}
    \vspace{0.5cm}
    \begin{tabular}{ |p{2cm}||p{3cm}|p{6cm}|}
    \hline
     \textbf{Method}& \textbf{Origin} &\textbf{Framework}\\
     \hline
     RIAD   & Industry & Reconstruction\\
     IterMask & Neuroimaging & Reconstruction\\
     ANDi & Neuroimaging & Reconstruction\\
     Disyre & Neuroimaging & Reconstruction + Synthetic Anomalies\\
     FAE & Neuroimaging & Feature-based with SSIM \\
     UniAD & Industry & Feature-based\\
     RD & Industry & Feature-based\\
     PatchCore & Industry & Feature-based\\
     \hline
    \end{tabular}
    \label{Table2}
\end{table*}

The \textit{reconstruction-based} approaches used in this study include Reconstruction-by-Inpainting Anomaly Detection (RIAD)\cite{zavrtanik2021reconstruction},  Iterative Spatial Mask-Refining (IterMask)\cite{liang2024itermask},  Aggregated Normative Diffusion (ANDi)\cite{frotscher2023unsupervised} and Diffusion-Inspired Synthetic Restoration (Disyre)\cite{naval2024ensembled}. We parameterized all \textit{reconstruction-based} approaches with the same U-Net architecture that was inspired by the DDPM++ model from \cite{song2020score}.

RIAD\cite{zavrtanik2021reconstruction} is a self-supervised approach trained to predict masked regions of varying sizes within an image. The image is first partitioned into regions of size $k\times k$, each of which is randomly divided into $n$ disjoint subsets that serve as the masking patterns. For each subset, the network predicts a reconstruction, and the $n$ reconstructions are combined to yield an image in which every pixel has been estimated by the model. This process is repeated multiple times while varying the parameter $k$, thereby enforcing predictions at multiple spatial scales. The reconstruction losses across scales are averaged using the multi-scale gradient magnitude similarity measure, which encourages fidelity to structural features. The key idea is to obscure potentially anomalous regions while enabling the network to generalize across diverse lesion types through complex multi-scale masking strategies. In our experiments, we increased $k$ to ensure coverage of large brain lesions while proportionally increasing $n$, thereby maintaining sufficient contextual information for accurate reconstruction.

IterMask\cite{liang2024itermask} is a self-supervised approach that utilizes two models trained with multiple masking strategies to predict the original image. One model is trained using the masking of low-frequency components in an image. The other model is trained by additionally using randomly generated masks in pixel space. Consequently, both models receive the high-frequency components of the image as input, while the second one is able to integrate the complete frequency information from random image locations for reconstruction. Both models are used to create an iterative process that starts by first predicting the image from its high-frequency components using the first model, and then iteratively applying the second model on the resulting mask from the previous step, aiming to shrink the mask towards the anomalies. At each iteration, a new mask is generated by thresholding the reconstruction error from the previous step. The threshold is determined on a healthy validation set, and the iterative process is terminated once the relative change between consecutive steps falls below a predefined ratio.

ANDi\cite{frotscher2023unsupervised} is a diffusion model\cite{ddpm,kingma2021variational,hoogeboom2023simple} trained using Gaussian pyramidal noise, which injects perturbations at multiple spatial scales and thereby enhances sensitivity to low-frequency structures. This training strategy allows the network to effectively reason about broad, slowly varying anomalies throughout the entire denoising trajectory, rather than focusing solely on high-frequency details in the early time steps. To compute anomaly evidence, ANDi evaluates a selected subset of diffusion time steps. At each step, the predicted mean of the Gaussian transition is compared against the image-conditioned mean, and the squared reconstruction error is used as the local deviation measure. This step-wise evaluation captures discrepancies that may emerge at different stages of the diffusion process. To derive a single anomaly map, deviations across timesteps are aggregated using the geometric mean, a choice that emphasizes consistent deviations across scales while attenuating spurious errors occurring at isolated steps. The resulting anomaly map thus integrates information across multiple diffusion time steps and frequency ranges, offering a principled estimate of abnormality that is both spatially localized and robust to noise.

Disyre\cite{naval2024ensembled} is inspired by diffusion models and employs synthetic anomalies to construct a corruption process. Anomalies are generated using a novel Disentangled Anomaly Generation (DAG) framework, which independently samples shape, texture, and intensity attributes from uniform distributions. The shape of each anomaly is created by randomly selecting cuboids, spheres, or other 3D primitives, which are further modified using affine transformations and smoothed with a Gaussian kernel to blend with the surrounding tissue. Texture is defined through Foreign Patch Interpolation (FPI), where a random patch from the training set is inserted into the target image using a convex combination with a randomly sampled interpolation factor. The patch is normalized before insertion to prevent confounding between texture and intensity. Intensity is determined by sampling a bias factor and a tissue class identified through k-means clustering, and then adjusting the intensities of the selected tissue within the anomaly mask according to the bias factor. In combination, these disentangled attributes govern the corruption process, and the network is trained in a diffusion-style framework to predict the original anomaly-free image, effectively learning to reverse the synthetic corruption.

\textit{Feature-based} approaches included in this study were Structural Feature-Autoencoders (FAE)\cite{meissen2022unsupervised},  Unified Model for Multi-class Anomaly Detection (UniAD)\cite{you2022unified}, Reverse Distillation (RD)\cite{deng2022anomaly} and PatchCore\cite{roth2022towards}. All \textit{feature-based} approaches were parameterized by their original network or partially modified to fit the image resolution used in the experiments.

FAE\cite{meissen2022unsupervised} is a convolutional autoencoder that operates on embeddings extracted from multiple layers of a ResNet pretrained on ImageNet. The network is trained to reconstruct these embeddings using the Structural Similarity Index Measure (SSIM) as the loss function. At inference, anomaly detection is performed by computing the SSIM between the input and reconstructed feature maps, followed by averaging the similarities to produce the final anomaly score.

UniAD\cite{you2022unified} is a transformer-based network that introduces specialized layer types for anomaly detection. They are introduced to prevent the ``identical shortcut'' that causes the network to simply copy the input. A key innovation is the neighbor masked attention layer, a variant of standard attention in which tokens from the local neighborhood are masked prior to the attention calculation. In addition, the architecture features a novel decoder that employs multiple query embeddings, which are fused with encoder representations and integrated with the output from the previous layer. In order for the network to learn the identity mapping, the query embeddings need to be sensitive to the input, thereby reducing the reconstruction ability for abnormal samples. Similar to the FAE, it leverages embeddings from an ImageNet-pretrained network, specifically EfficientNet; however, unlike the FAE, Gaussian noise is added to the input features, forcing the network to jointly learn reconstruction and denoising. The model is trained with an L2 loss. At inference, anomalies are detected by computing the Euclidean distance between the input and reconstructed pixel-level features.

RD\cite{deng2022anomaly} applies a knowledge distillation framework to anomaly detection. A pre-trained ImageNet encoder serves as the teacher network, while a trainable bottleneck embedding module maps the teacher’s representations into a more compact code. In conjunction, a decoder is trained to reconstruct the teacher’s embeddings from this bottleneck representation, with training guided by maximizing cosine similarity between the reconstructed outputs and the teacher’s original embeddings. The key idea is that, due to the heterogeneous architecture and the compression introduced by the bottleneck, the student network cannot perfectly replicate the teacher’s embeddings for novel or anomalous data. As a result, anomalies can be identified at test time by measuring the cosine similarity between the pixel-level features of the input and output, with lower similarity indicating abnormality.

PatchCore\cite{roth2022towards} employs a memory bank of preprocessed embeddings for anomaly detection. A WideResNet-50 pretrained on ImageNet is used as the backbone, from which embeddings are extracted from intermediate layers. Prior to assembling the memory bank, the features are refined to enlarge the receptive field: overlapping patches are sampled from the feature maps, and adaptive average pooling is applied to aggregate information from each neighborhood into a single feature vector. The resulting representations form the memory bank of nominal features. To improve efficiency at inference, the memory bank is downsampled using greedy coreset subsampling. Anomaly detection is then performed by computing the mean L2-distance to the nearest entries in the memory bank. In our MRI experiments, we observed that performance improved when considering only the distance to the single closest entry, rather than averaging across multiple neighbors.

All models were trained for a maximum of 100,000 gradient update steps, and checkpoints were saved after every 5000 steps. Then, the best-performing checkpoint was chosen from the validation dataset for each method. We noticed that many methods reached optimal performance on the validation dataset before convergence. 

All code, along with our adaptations of the evaluated methods into reproducible and testable workflows, is available on GitHub (\url{https://github.com/AlexanderFrotscher/UAD-IMAG}) and has been forked to the MHMlab repository (\url{https://github.com/MHM-lab}). Prior to submission, we contacted the lead authors of the original papers and invited them to review our implementations via the shared GitHub repository. This process was designed to ensure transparency, fairness, and author-validated benchmarking.

\subsection{Post-Hoc Statistics and Model Evaluation}
\label{Sec:Post}
After obtaining the anomaly maps, three-dimensional median filtering with a kernel size of four is applied to all algorithms. Threshold selection has been performed on the original anomaly maps and has been transferred to the median-filtered ones. For evaluating the anomaly map, many different metrics can be used; that is, all metrics that can be used for binary classification can, in principle, work for the evaluation of the UAD methods. In the medical community, two of these are now widely accepted as the standards: the Dice score and the area under the precision-recall curve (AUPRC). Note that for brain MRI, the area under the receiver operating characteristic curve is less important due to the high class imbalance between positives and negatives. Here, we mainly report the Dice score and present all AUPRC values in the Supplementary Information. Table 1 - 4. All threshold-dependent metrics have been calculated on a per-individual basis and averaged across individuals when reporting mean values, whereas the AUPRC has been calculated on the complete dataset once. To evaluate the influence of the scanner (the specific MRI device) and sex effects, the distributions of the performance measures and groups have been analyzed using the Mann-Whitney U test. It is a rank-based test that does not assume a specific parametric distribution and can indicate significant differences for all aspects of the distribution, e.g., location, scale, and shape. Multiple testing corrections using the Benjamini-Hochberg method have been applied to correct for the four datasets corresponding to the different lesion loads. Furthermore, age effects have been analyzed using the Spearman rank correlation test to assess nonlinear correlations between performance and age. For all statistical tests, a significance level $\alpha=0.05$ has been used. To analyze the magnitude of the sex effects, Cohen’s~\textit{d} was calculated with the male group as the first group. Therefore, all positive Cohen’s~\textit{d} values indicate that the FPRs were higher for the male group.

\section{Acknowledgements}
AF and TW thank the International Max Planck Research School for Intelligent Systems (IMPRS-IS) for their support. TW acknowledges funding from the German Research Foundation (DFG) Emmy Noether: 513851350 and the BMBF/DLR Project FEDORA: 01EQ2403G. This work was supported by the BMBF-funded de.NBI Cloud within the German Network for Bioinformatics Infrastructure (de.NBI) (031A532B, 031A533A, 031A533B, 031A534A, 031A535A, 031A537A, 031A537B, 031A537C, 031A537D, 031A538A).

\bibliographystyle{plainnat}
\bibliography{references.bib}

\onecolumn
\appendix
\include{supplementary}
\end{document}

%% file: supplementary.tex
\section{Supplementary Materials}

\subsection{Algorithmic Performance Across Brain Lesions}
Here, we report the results of the main analysis in tabular form for the BraTS20 dataset (see \cref{Tab:BraTS}), the ATLAS dataset (see \cref{Tab:ATLAS}), the Shifts dataset (see \cref{Tab:Shifts}), and the WMH dataset (see \cref{Tab:WMH}). The threshold dependent metrics $\lceil \text{Dice} \rceil$ and $\text{Dice}_{\text{Estimate}}$ are calculated per individual, and the average is reported. AUPRC is calculated on the complete dataset once.
\begin{table*}[ht]
\centering
\caption{Anomaly detection performance measured in AUPRC, $\lceil \text{Dice} \rceil$ and $\text{Dice}_{\text{Estimate}}$ on the BraTS20 test dataset (332 subjects). We report the results without post-processing and when using three dimensional median filtering (MF) with a kernel size of four. In the main paper only the MF values are used in the plots.}
\label{Tab:BraTS}
\renewcommand{\arraystretch}{0.5}
\resizebox{1.0\textwidth}{!}{\begin{tabular}{l c c c c c c}
 & \multicolumn{3}{c}{BraTS20 T1w}& \multicolumn{3}{c}{BraTS20 T2w}\\ 
\cmidrule(lr){2-4}
\cmidrule(lr){5-7}
Method &
AUPRC& $\lceil \text{Dice} \rceil$ &$\text{Dice}_{\text{Estimate}}$
&
AUPRC& $\lceil \text{Dice} \rceil$ &$\text{Dice}_{\text{Estimate}}$\\
\midrule
RIAD
&0.100&0.148&0.144&0.148&0.203&0.203\\
\textcolor{gray}{w/ MF}
&0.120&0.161&0.161&0.187&0.236&0.236\\
\midrule
IterMask
&0.294&0.293&0.290&0.250&0.256&0.256\\
\textcolor{gray}{w/ MF}
&0.308&0.301&0.297&0.252&0.258&0.257\\
\midrule
ANDi
&0.177&0.216&0.197&0.410&0.359&0.358\\
\textcolor{gray}{w/ MF}
&0.264&0.265&0.232&0.518&0.453&0.425\\
\midrule
Disyre
&0.158&0.258&0.249&0.638&0.594&0.594\\
\textcolor{gray}{w/ MF}
&0.178&0.274&0.264&0.701&0.646&0.645\\
\midrule
FAE
&0.460&0.427&0.427&0.505&0.457&0.457\\
\textcolor{gray}{w/ MF}
&0.480&0.437&0.437&0.522&0.467&0.467\\
\midrule
UniAD
&0.180&0.227&0.227&0.352&0.361&0.361\\
\textcolor{gray}{w/ MF}
&0.198&0.239&0.239&0.386&0.381&0.381\\
\midrule
RD
&0.143&0.221&0.221&0.222&0.272&0.272\\
\textcolor{gray}{w/ MF}
&0.154&0.231&0.231&0.247&0.283&0.283\\
\midrule
PatchCore
&0.223&0.327&0.314&0.347&0.366&0.366\\
\textcolor{gray}{w/ MF}
&0.226&0.334&0.321&0.355&0.372&0.372\\
\bottomrule
\end{tabular}}
\end{table*}

\begin{table*}[ht]
\centering
\caption{Anomaly detection performance measured in AUPRC, $\lceil \text{Dice} \rceil$ and $\text{Dice}_{\text{Estimate}}$ on the T1w images of the ATLAS test dataset (ATLAS-I 243 individuals, ATLAS-O 212 individuals, ATLAS-N 159 individuals). We report the results without post-processing and when using three dimensional median filtering (MF) with a kernel size of four. In the main paper only the MF values are used in the plots.}
\label{Tab:ATLAS}
\renewcommand{\arraystretch}{1.2}
\resizebox{1.0\textwidth}{!}{\begin{tabular}{l c c c c c c c c c}
 & \multicolumn{3}{c}{ATLAS-I}  & \multicolumn{3}{c}{ATLAS-O}  & \multicolumn{3}{c}{ATLAS-N}\\ 
\cmidrule(lr){2-4} \cmidrule(lr){5-7} \cmidrule(lr){8-10}
Method &
AUPRC& $\lceil \text{Dice} \rceil$ &$\text{Dice}_{\text{Estimate}}$
&
AUPRC& $\lceil \text{Dice} \rceil$ &$\text{Dice}_{\text{Estimate}}$
&
AUPRC& $\lceil \text{Dice} \rceil$ &$\text{Dice}_{\text{Estimate}}$\\
\midrule
RIAD
&0.041&0.090&0.064&0.066&0.089&0.076&0.009&0.025&0.013
\\
\textcolor{gray}{w/ MF}
&0.050&0.129&0.081&0.105&0.126&0.097&0.015&0.045&0.017
\\
\midrule
IterMask
&0.037&0.116&0.108&0.093&0.090&0.089&0.055&0.041&0.020
\\
\textcolor{gray}{w/ MF}
&0.037&0.118&0.113&0.096&0.091&0.090&0.063&0.041&0.021
\\
\midrule
ANDi
&0.222&0.188&0.183&0.119&0.143&0.138&0.033&0.074&0.042
\\
\textcolor{gray}{w/ MF}
&0.333&0.285&0.281&0.188&0.209&0.209&0.118&0.145&0.118
\\
\midrule
Disyre
&0.590&0.408&0.390&0.450&0.351&0.348&0.236&0.170&0.152
\\
\textcolor{gray}{w/ MF}
&0.616&0.434&0.432&0.500&0.386&0.384&0.303&0.196&0.191
\\
\midrule
FAE
&0.271&0.177&0.177&0.106&0.161&0.161&0.059&0.048&0.044
\\
\textcolor{gray}{w/ MF}
&0.287&0.183&0.183&0.110&0.167&0.166&0.062&0.049&0.047
\\
\midrule
UniAD
&0.214&0.173&0.162&0.115&0.162&0.162&0.043&0.040&0.030
\\
\textcolor{gray}{w/ MF}
&0.240&0.188&0.178&0.125&0.178&0.178&0.049&0.045&0.035
\\
\midrule
RD
&0.105&0.113&0.109&0.083&0.122&0.121&0.029&0.028&0.024
\\

\textcolor{gray}{w/ MF}
&0.119&0.122&0.117&0.090&0.130&0.130&0.041&0.032&0.026
\\
\midrule
PatchCore
&0.312&0.181&0.148&0.127&0.160&0.147&0.116&0.046&0.022
\\
\textcolor{gray}{w/ MF}
&0.322&0.186&0.148&0.131&0.165&0.149&0.124&0.047&0.021
\\
\bottomrule
\end{tabular}}
\end{table*}

\begin{table*}[ht]
\centering
\caption{Anomaly detection performance measured in AUPRC, $\lceil \text{Dice} \rceil$ and $\text{Dice}_{\text{Estimate}}$ on the Shifts test dataset (68 individuals). We report the results without post-processing and when using three dimensional median filtering (MF) with a kernel size of four. In the main paper only the MF values are used in the plots.}
\label{Tab:Shifts}
\renewcommand{\arraystretch}{0.5}
\resizebox{1.0\textwidth}{!}{\begin{tabular}{l c c c c c c}
 & \multicolumn{3}{c}{Shifts T1w}  & \multicolumn{3}{c}{Shifts T2w}\\ 
\cmidrule(lr){2-4}
\cmidrule(lr){5-7}
Method &
AUPRC& $\lceil \text{Dice} \rceil$ &$\text{Dice}_{\text{Estimate}}$
&
AUPRC& $\lceil \text{Dice} \rceil$ &$\text{Dice}_{\text{Estimate}}$\\
\midrule
RIAD
&0.021&0.050&0.038&0.020&0.051&0.040\\
\textcolor{gray}{w/ MF}
&0.024&0.062&0.045&0.021&0.064&0.050\\
\midrule
IterMask
&0.016&0.044&0.034&0.010&0.020&0.015\\
\textcolor{gray}{w/ MF}
&0.017&0.045&0.031&0.010&0.021&0.015\\
\midrule
ANDi
&0.072&0.123&0.123&0.054&0.099&0.099\\
\textcolor{gray}{w/ MF}
&0.110&0.143&0.125&0.061&0.128&0.104\\
\midrule
Disyre
&0.148&0.151&0.148&0.235&0.208&0.197\\
\textcolor{gray}{w/ MF}
&0.166&0.156&0.139&0.234&0.205&0.172\\
\midrule
FAE
&0.022&0.032&0.017&0.043&0.049&0.032\\
\textcolor{gray}{w/ MF}
&0.022&0.032&0.015&0.044&0.048&0.031\\
\midrule
UniAD
&0.030&0.045&0.045&0.04&0.059&0.053\\
\textcolor{gray}{w/ MF}
&0.033&0.048&0.048&0.046&0.063&0.055\\
\midrule
RD
&0.029&0.028&0.024&0.032&0.043&0.035\\
\textcolor{gray}{w/ MF}
&0.041&0.032&0.026&0.034&0.045&0.035\\
\midrule
PatchCore
&0.043&0.057&0.026&0.045&0.056&0.017\\
\textcolor{gray}{w/ MF}
&0.046&0.058&0.025&0.049&0.058&0.015\\
\bottomrule
\end{tabular}}
\end{table*}

\begin{table*}[ht]
\centering
\caption{Anomaly detection performance measured in AUPRC, $\lceil \text{Dice} \rceil$ and $\text{Dice}_{\text{Estimate}}$ on the T1w images of the WMH test dataset (54 individuals). We report the results without post-processing and when using three dimensional median filtering (MF) with a kernel size of four. In the main paper only the MF values are used in the plots.}
\label{Tab:WMH}
\renewcommand{\arraystretch}{0.5}
\resizebox{0.6\textwidth}{!}{\begin{tabular}{l c c c}
 & \multicolumn{3}{c}{WMH}\\ 
\cmidrule(lr){2-4}
Method &
AUPRC& $\lceil \text{Dice} \rceil$ &$\text{Dice}_{\text{Estimate}}$\\
\midrule
RIAD
&0.019&0.037&0.035\\
\textcolor{gray}{w/ MF}
&0.023&0.044&0.042\\
\midrule
IterMask
&0.019&0.037&0.036\\
\textcolor{gray}{w/ MF}
&0.021&0.037&0.036\\
\midrule
ANDi
&0.028&0.053&0.051\\
\textcolor{gray}{w/ MF}
&0.031&0.062&0.052\\
\midrule
Disyre
&0.085&0.119&0.112\\
\textcolor{gray}{w/ MF}
&0.098&0.130&0.109\\
\midrule
FAE
&0.019&0.032&0.018\\
\textcolor{gray}{w/ MF}
&0.019&0.032&0.016\\
\midrule
UniAD
&0.027&0.042&0.042\\
\textcolor{gray}{w/ MF}
&0.028&0.044&0.044\\
\midrule
RD
&0.022&0.038&0.028\\
\textcolor{gray}{w/ MF}
&0.023&0.039&0.027\\
\midrule
PatchCore
&0.044&0.072&0.050\\
\textcolor{gray}{w/ MF}
&0.045&0.074&0.049\\
\bottomrule
\end{tabular}}
\end{table*}

\subsection{Algorithmic Performance Across Healthy Individuals}
\begin{figure}[ht]
    \centering
    \includegraphics[scale=0.4]{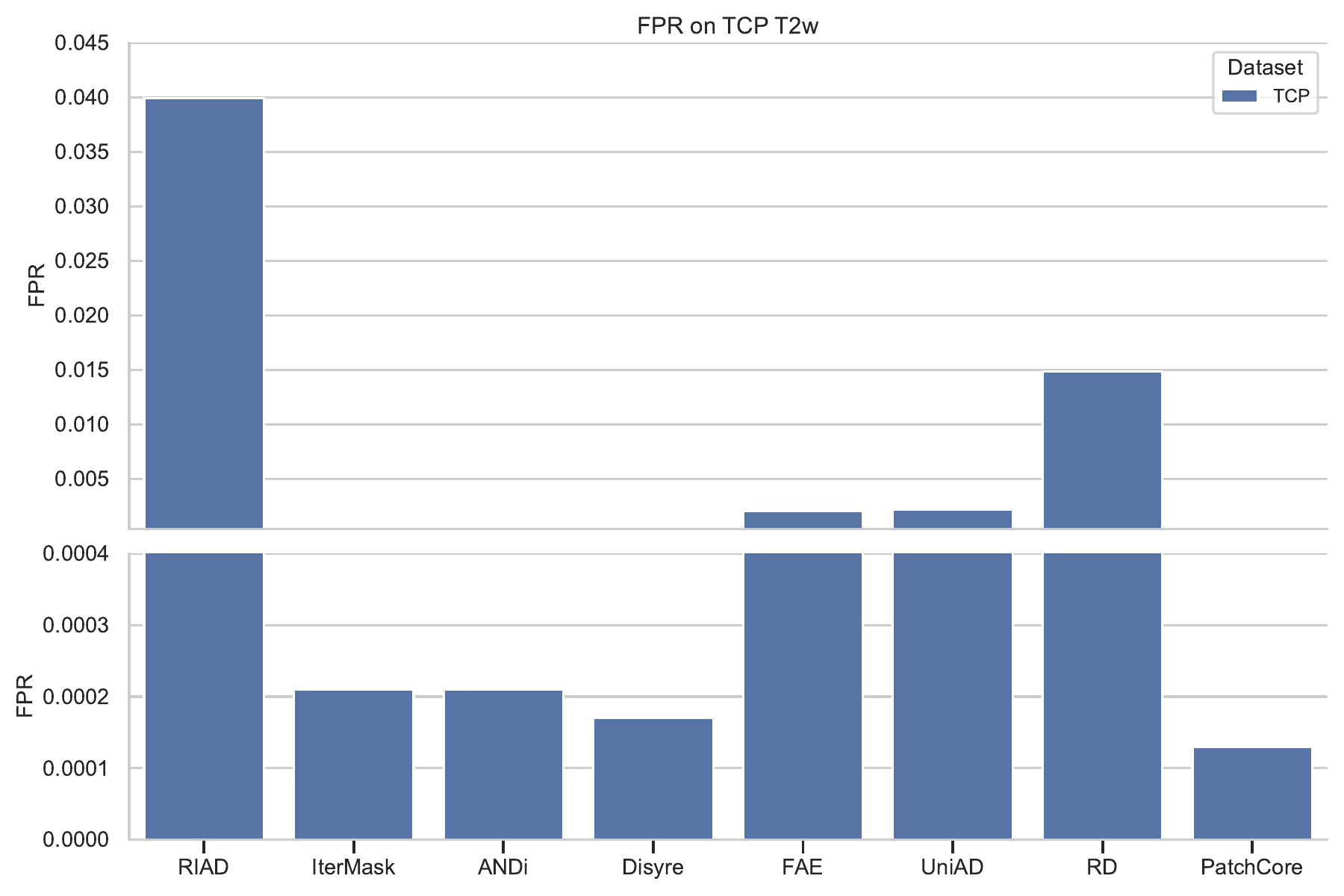}
    \caption{\textbf{False positive rates on the T2w healthy portion of the TCP datasets for all tested algorithms based on the \textit{estimated} threshold.}}
    \label{fig:Healthy-T2}
\end{figure}

The false positive rates for all datasets can be found in \cref{Tab:FPR} as well as the bar plot for the T2w FPRs in \cref{fig:Healthy-T2}

\begin{table*}[ht]
\centering
\caption{False positive rates on all datasets (FCON1T 139 individuals, FCON3T 105 individuals, TCP 92 individuals). We report the results without post-processing and when using three dimensional median filtering (MF) with a kernel size of four. In the main paper only the MF values are used in the plots.}
\label{Tab:FPR}
\renewcommand{\arraystretch}{1.0}
\resizebox{0.9\textwidth}{!}{\begin{tabular}{l c c c c}
& \multicolumn{4}{c}{Datasets}\\
\cmidrule(lr){2-5}
Method &
FCON1T& FCON3T & TCP T1w & TCP T2w\\
\midrule
RIAD
&$2.00 \times 10^{-1}$&$1.74 \times 10^{-1}$&$1.46 \times 10^{-1}$&$8.61 \times 10^{-2}$
\\
\textcolor{gray}{w/ MF}
&$1.23 \times 10^{-1}$&$1.06 \times 10^{-1}$&$8.30 \times 10^{-2}$&$4.00 \times 10^{-2}$
\\
\midrule
IterMask
&$1.67 \times 10^{-1}$&$1.50 \times 10^{-1}$&$2.15 \times 10^{-2}$&$5.50 \times 10^{-4}$
\\
\textcolor{gray}{w/ MF}
&$1.60 \times 10^{-1}$&$1.42 \times 10^{-1}$&$1.71 \times 10^{-2}$&$2.10 \times 10^{-4}$
\\
\midrule
ANDi
&$3.71 \times 10^{-2}$&$2.80 \times 10^{-2}$&$1.18 \times 10^{-2}$&$8.30 \times 10^{-3}$
\\
\textcolor{gray}{w/ MF}
&$4.82 \times 10^{-3}$&$5.83 \times 10^{-3}$&$3.79 \times 10^{-3}$&$2.10 \times 10^{-4}$
\\
\midrule
Disyre
&$2.56 \times 10^{-3}$&$5.00 \times 10^{-3}$&$1.69 \times 10^{-3}$&$1.08 \times 10^{-3}$
\\
\textcolor{gray}{w/ MF}
&$8.20 \times 10^{-4}$&$1.99 \times 10^{-3}$&$4.00 \times 10^{-4}$&$1.70 \times 10^{-4}$
\\
\midrule
FAE
&$1.77 \times 10^{-2}$&$1.49 \times 10^{-2}$&$3.01 \times 10^{-3}$&$2.67 \times 10^{-3}$
\\
\textcolor{gray}{w/ MF}
&$1.44 \times 10^{-2}$&$1.23 \times 10^{-2}$&$2.23 \times 10^{-3}$&$2.04 \times 10^{-3}$
\\
\midrule
UniAD
&$1.65 \times 10^{-2}$&$1.51 \times 10^{-2}$&$1.04 \times 10^{-2}$&$4.41 \times 10^{-3}$
\\
\textcolor{gray}{w/ MF}
&$1.12 \times 10^{-2}$&$1.01 \times 10^{-2}$&$6.64 \times 10^{-3}$&$2.17 \times 10^{-3}$
\\
\midrule
RD
&$9.12 \times 10^{-2}$&$3.91 \times 10^{-2}$&$1.77 \times 10^{-2}$&$1.93 \times 10^{-2}$
\\
\textcolor{gray}{w/ MF}
&$7.95 \times 10^{-2}$&$3.16 \times 10^{-2}$&$1.29 \times 10^{-2}$&$1.49 \times 10^{-2}$
\\
\midrule
PatchCore
&$2.09 \times 10^{-2}$&$5.56 \times 10^{-3}$&$4.20 \times 10^{-4}$&$2.70 \times 10^{-4}$
\\
\textcolor{gray}{w/ MF}
&$1.84 \times 10^{-2}$&$4.69 \times 10^{-3}$&$3.20 \times 10^{-4}$&$1.30 \times 10^{-4}$
\\
\bottomrule
\end{tabular}}
\end{table*}

\subsection{Impact of Domain Shifts and Lesion Load Variability}

Here we report the p-values for the different Mann-Whitney U tests before and after multiple testing correction. They can be found in \cref{Tab:p-value}. 

\begin{table}[t]
    \centering
    \caption{P-value, and corrected p-value using the Benjamini–Hochberg method for the Mann-Whitney U tests between Dice scores for ID and OOD pairs. We constructed eight datasets out of the ATLAS-I and ATLAS-O datasets corresponding to four matched ID–OOD pairs, stratified by lesion load percentiles derived from the full ATLAS cohort. We group volumes above the 75th percentile (Top), above the 50th percentile to 75th (Upper), above the 25th percentile to 50th percentile (Middle), and below the 25th percentile (Lower). The reported values are derived from median filtered anomaly maps and the \textit{estimated} threshold.}
    \label{Tab:p-value}
    \renewcommand{\arraystretch}{1.0}
     \resizebox{1.0\textwidth}{!}{\begin{tabular}{|p{2cm}|p{2cm}|p{2cm}|p{2cm}|p{2cm}|}
    \hline
     \textbf{Algorithm} &\textbf{Top} &\textbf{Upper} &\textbf{Middle} &\textbf{Lower}\\
     \hline
     \multirow{8}{*}{\rotatebox[origin=c]{90}{\textit{\footnotesize standard p-value~~}}}
     RIAD&$5.12 \times 10^{-1}$&$9.42 \times 10^{-2}$&$3.14\times 10^{-1}$&$2.49\times 10^{-1}$\\
     ~~~IterMask&$5.89 \times 10^{-3}$&$4.81\times 10^{-2}$&$8.96\times 10^{-2}$&$1.77\times 10^{-2}$\\
     ~~~ANDi&$1.88 \times 10^{-8}$&$1.44\times 10^{-1}$&$1.21\times 10^{-1}$&$1.60\times 10^{-1}$\\
     ~~~Disyre&$4.48 \times 10^{-9}$&$2.35\times 10^{-2}$&$5.28\times 10^{-1}$&$7.25 \times 10^{-2}$\\
     ~~~FAE&$5.59 \times 10^{-2}$&$1.35\times 10^{-2}$&$1.72\times 10^{-2}$&$7.13\times 10^{-1}$\\
     ~~~UniAD&$1.02 \times 10^{-1}$&$3.06\times 10^{-1}$&$2.16\times 10^{-1}$&$4.86\times 10^{-1}$\\
     ~~~RD&$3.71 \times 10^{-1}$&$1.35\times 10^{-1}$&$5.59 \times 10^{-2}$&$4.98\times 10^{-1}$\\
     ~~~PatchCore&$8.96 \times 10^{-1}$&$5.06\times 10^{-2}$&$6.07\times 10^{-1}$&$4.91\times 10^{-1}$\\
     \hline
     \multirow{8}{*}{\rotatebox[origin=c]{90}{\textit{\footnotesize corrected p-value~~}}}
     RIAD&$5.12 \times 10^{-1}$&$3.77 \times 10^{-1}$&$4.19\times 10^{-1}$&$4.19\times 10^{-1}$\\
     ~~~IterMask&$2.35 \times 10^{-2}$&$6.42\times 10^{-2}$&$8.96\times 10^{-2}$&$3.54\times 10^{-2}$\\
     ~~~ANDi&$7.51 \times 10^{-8}$&$1.60\times 10^{-1}$&$1.60\times 10^{-1}$&$1.60\times 10^{-1}$\\
     ~~~Disyre&$1.79 \times 10^{-8}$&$4.70\times 10^{-2}$&$5.28\times 10^{-1}$&$9.67 \times 10^{-2}$\\
     ~~~FAE&$7.45 \times 10^{-2}$&$3.45\times 10^{-2}$&$3.45\times 10^{-2}$&$7.13\times 10^{-1}$\\
     ~~~UniAD&$4.08 \times 10^{-1}$&$4.08\times 10^{-1}$&$4.08\times 10^{-1}$&$4.86\times 10^{-1}$\\
     ~~~RD&$4.95 \times 10^{-1}$&$2.69\times 10^{-1}$&$2.24 \times 10^{-1}$&$4.98\times 10^{-1}$\\
     ~~~PatchCore&$8.96 \times 10^{-1}$&$2.03\times 10^{-1}$&$8.09\times 10^{-1}$&$8.09\times 10^{-1}$\\
     \hline
    \end{tabular}}
\end{table}

\begin{table}[t]
    \centering
    \caption{P-values for the independent variables and the interaction term as well as the degrees of freedom (site, size, interaction) used for the Chi-Square approximation for the Scheirer–Ray–Hare tests. The Dice scores for each method were selected as the dependent variable and the lesion load categories and the ID-OOD dummy variable as the independent variables. The reported values are derived from median filtered anomaly maps and the \textit{estimated} threshold.}
    \label{Tab:Hare}
    \renewcommand{\arraystretch}{1.0}
     \resizebox{1.0\textwidth}{!}{\begin{tabular}{|p{2cm} |p{2.5cm}|p{3.5cm}|p{3.5cm}|p{3.5cm}|}
    \hline
     \textbf{Algorithm} &\textbf{p-value Site} &\textbf{p-value Lesion Size} &\textbf{p-value Interaction} &\textbf{Degrees of Freedom}\\
     \hline
     RIAD&$7.10 \times 10^{-1}$&$0.0 \times 10^{0}$&$9.68 \times 10^{-1}$ & $1, 3, 3$\\
     IterMask&$5.99 \times 10^{-1}$&$0.0 \times 10^{0}$&$3.58 \times 10^{-1}$ & $1, 3, 3$\\
     ANDi&$4.68 \times 10^{-5}$&$0.0 \times 10^{0}$&$2.05 \times 10^{-1}$ & $1, 3, 3$\\
     Disyre&$9.93 \times 10^{-4}$&$0.0 \times 10^{0}$&$9.99 \times 10^{-3}$ & $1, 3, 3$\\
     FAE&$5.36 \times 10^{-2}$&$0.0 \times 10^{0}$&$7.55 \times 10^{-1}$ & $1, 3, 3$\\
     UniAD&$1.77 \times 10^{-1}$&$0.0 \times 10^{0}$&$9.78 \times 10^{-1}$ & $1, 3, 3$\\
     RD&$3.94 \times 10^{-1}$&$0.0 \times 10^{0}$&$8.95 \times 10^{-1}$ & $1, 3, 3$\\
     PatchCore&$5.15 \times 10^{-1}$&$0.0 \times 10^{0}$&$5.91 \times 10^{-1}$ & $1, 3, 3$\\
     \hline
     
    \end{tabular}}
\end{table}

\subsection{Impact of demographics on false positive rates and lesion identification.}

\begin{figure*}[ht]
    \centering
    \includegraphics[scale=0.73]{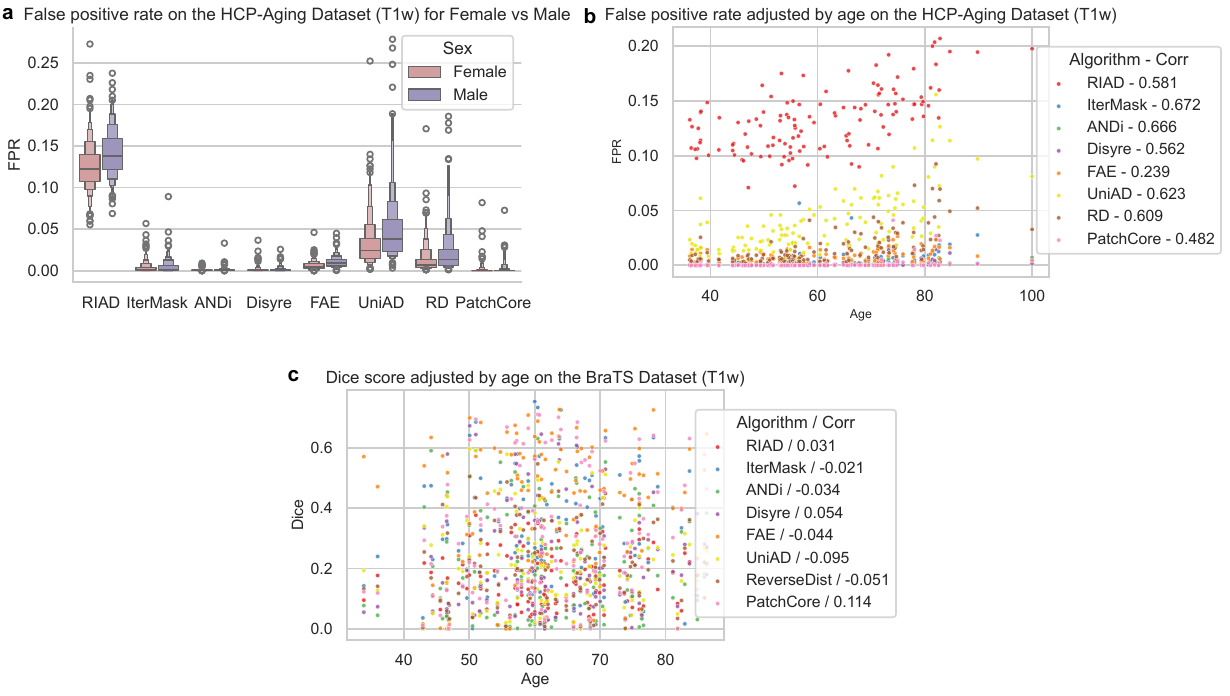}
    \caption{\textbf{Impact of demographics on false positive rates and lesion identification for T1w images. a)} The false positive rate on the HCP Aging dataset for the female and male group. All groups show significant differences based on the Mann-Whitney U test. Significance level $\alpha=0.05$ was used for all tests. \textbf{b)} The false positive rate on the HCP Aging dataset adjusted by age. The Spearman rank correlation test displayed that all algorithms show significant positive correlation indicating that older subjects are more likely to have an anomaly assigned to a voxel that is normal.  \textbf{c)} The Dice score on the BraTS dataset adjusted by age.}
    \label{fig:demoT1w}
\end{figure*}

Here we show the results for T1w images of the studied impact of inter-individual anatomical variability linked to demographic attributes such as age and sex in the HCP Aging and BraTS datasets. All additional plots can be found in \cref{fig:demoT1w} and all results in numeric format can be found in \cref{Tab:HCPAging}.

\begin{table}[t]
    \centering
    \caption{P-value, mean, standard deviation and Cohen's~\textit{d} of FPR values calculated on HCP Aging for all algorithms. The reported values are derived from median filtered anomaly maps.}
    \label{Tab:HCPAging}
    \renewcommand{\arraystretch}{1.3}
     \resizebox{1.0\textwidth}{!}{\begin{tabular}{|p{2cm}| p{2cm}|p{2cm}|p{2.3cm}|p{2cm}|p{2cm}|p{2cm}|}
    \hline
     \textbf{Algorithm} &\textbf{p-value} &\textbf{Mean male} &\textbf{Mean female} &\textbf{Std male} &\textbf{Std female} &\textbf{Cohen's~\textit{d}}\\
     \hline
     \multirow{8}{*}{\rotatebox[origin=c]{90}{\textit{\footnotesize T1w images~~}}}
     RIAD&$5.88 \times 10^{-15}$&$1.41 \times 10^{-1}$&$1.25\times 10^{-1}$&$2.78\times 10^{-2}$&$2.79\times 10^{-2}$&$5.81 \times 10^{-1}$\\
     ~~~IterMask&$1.99 \times 10^{-4}$&$4.87\times 10^{-3}$&$3.34\times 10^{-3}$&$8.40\times 10^{-3}$&$6.43\times 10^{-3}$&$2.08\times 10^{-1}$\\
     ~~~ANDi&$1.10 \times 10^{-5}$&$1.15\times 10^{-3}$&$7.66\times 10^{-4}$&$2.26\times 10^{-3}$&$1.08\times 10^{-3}$&$2.24\times 10^{-1}$\\
     ~~~Disyre&$1.84 \times 10^{-8}$&$1.38\times 10^{-3}$&$1.01\times 10^{-3}$&$2.76 \times 10^{-3}$&$2.84\times 10^{-3}$&$1.33\times 10^{-1}$\\
     ~~~FAE&$1.19 \times 10^{-25}$&$1.05\times 10^{-2}$&$5.97\times 10^{-3}$&$6.72\times 10^{-3}$&$5.15\times 10^{-3}$&$7.66\times 10^{-1}$\\
     ~~~UniAD&$4.19 \times 10^{-17}$&$4.95\times 10^{-2}$&$3.16\times 10^{-2}$&$4.02\times 10^{-2}$&$2.69\times 10^{-2}$&$5.37\times 10^{-1}$\\
     ~~~RD&$2.53 \times 10^{-14}$&$2.20\times 10^{-2}$&$1.24 \times 10^{-2}$&$2.66\times 10^{-2}$&$1.57\times 10^{-2}$&$4.54\times 10^{-1}$\\
     ~~~PatchCore&$1.50 \times 10^{-4}$&$1.66\times 10^{-3}$&$1.10\times 10^{-3}$&$6.10\times 10^{-3}$&$5.65\times 10^{-3}$&$9.54\times 10^{-2}$\\
     \hline
     \multirow{8}{*}{\rotatebox[origin=c]{90}{\textit{\footnotesize T2w images~~}}}
     RIAD&$4.15 \times 10^{-7}$&$6.70 \times 10^{-2}$&$5.76\times 10^{-2}$&$2.81\times 10^{-2}$&$2.66\times 10^{-2}$&$3.41 \times 10^{-1}$\\
     ~~~IterMask&$7.91 \times 10^{-3}$&$1.53\times 10^{-3}$&$9.98\times 10^{-4}$&$3.63\times 10^{-3}$&$2.24\times 10^{-3}$&$1.82\times 10^{-1}$\\
     ~~~ANDi&$2.49 \times 10^{-3}$&$6.85\times 10^{-4}$&$5.80\times 10^{-4}$&$1.36\times 10^{-3}$&$1.06\times 10^{-3}$&$8.78\times 10^{-2}$\\
     ~~~Disyre&$4.66 \times 10^{-6}$&$5.35\times 10^{-4}$&$3.40\times 10^{-4}$&$1.49 \times 10^{-3}$&$7.66\times 10^{-4}$&$1.70\times 10^{-1}$\\
     ~~~FAE&$1.02 \times 10^{-10}$&$6.37\times 10^{-3}$&$4.41\times 10^{-3}$&$8.52\times 10^{-3}$&$7.06\times 10^{-3}$&$2.53\times 10^{-1}$\\
     ~~~UniAD&$1.29 \times 10^{-10}$&$9.57\times 10^{-3}$&$5.72\times 10^{-3}$&$1.33\times 10^{-2}$&$7.69\times 10^{-3}$&$3.66\times 10^{-1}$\\
     ~~~RD&$5.79 \times 10^{-8}$&$7.80\times 10^{-3}$&$4.82 \times 10^{-3}$&$1.17\times 10^{-2}$&$7.54\times 10^{-2}$&$3.10\times 10^{-1}$\\
     ~~~PatchCore&$1.54 \times 10^{-4}$&$1.61\times 10^{-3}$&$8.61\times 10^{-4}$&$5.74\times 10^{-3}$&$3.84\times 10^{-3}$&$1.57\times 10^{-1}$\\
     \hline
    \end{tabular}}
\end{table}

\subsection{Influence of Network choice on Feature-based approaches}
\begin{figure*}[ht]
    \centering
    \includegraphics[scale=0.6]{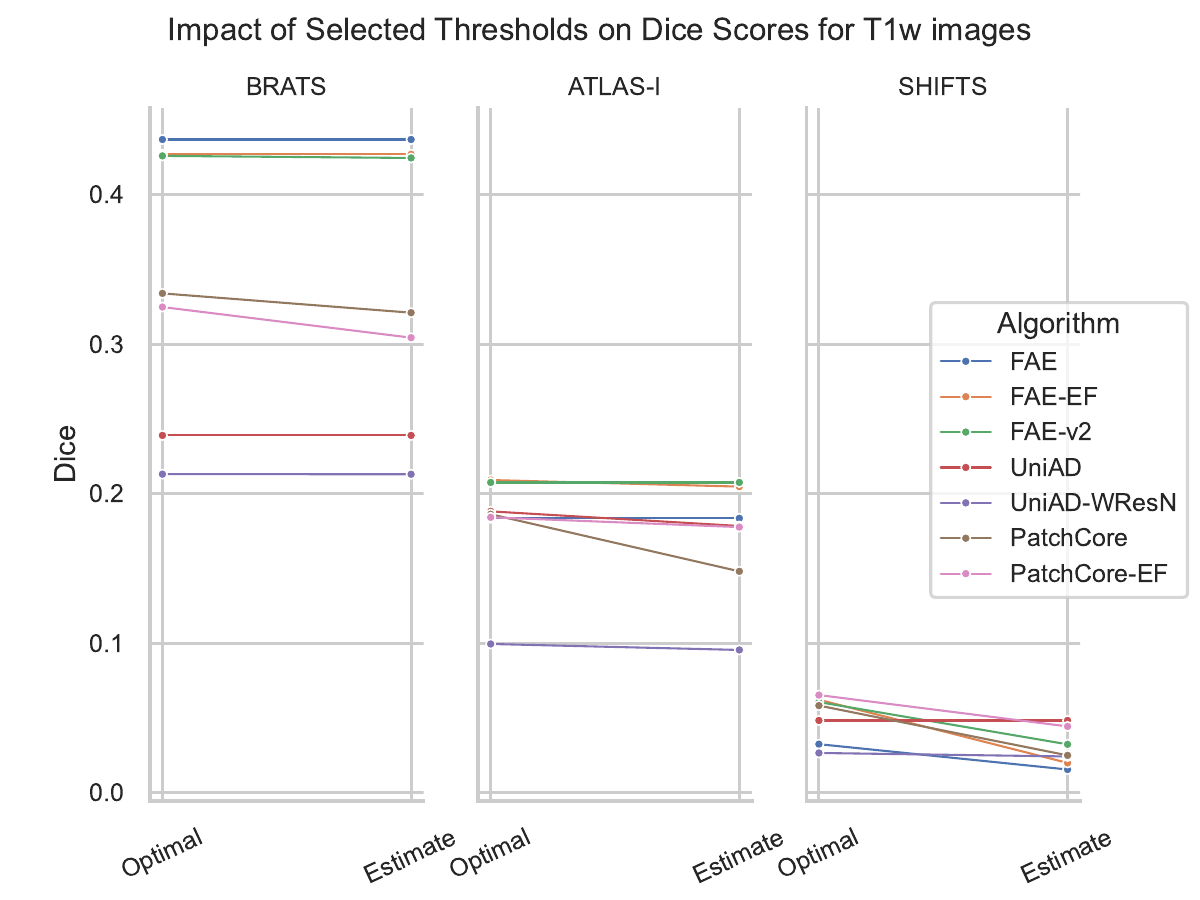}
    \caption{\textbf{Impact of the pretrained network on state-of-the-art \textit{feature-based} unsupervised anomaly detection.} The performance of the algorithms on T1w images were reported using two different thresholds, the \textit{optimal} threshold, defined as the maximum possible Dice score optimized in the test set, thus potentially susceptible to bias but standard in the field. Second, the \textit{estimated} threshold optimized on the validation data set, then fixed and performance for that threshold reported on the untouched test set, thus unbiased but not the standard in the field. Here we tested multiple \textit{feature-based} approaches while altering the pretrained network.}
    \label{fig:feature}
\end{figure*}
To further test the performance of the \textit{feature-based} approaches, different pretrained networks that were trained on the classification task of ImageNet have been selected. The models and pretrained weights from torchvision have been used for all experiments. We observed that version one of the set of weights performed better for all anomaly detection tasks tested and that ImageNet specific normalization for mean and standard deviation on the images yielded slightly reduced performance. In \cref{fig:feature} we additionally show FAE and PatchCore with features extracted from EfficientNet-B4 and UniAD when using features from WideResNet50. Furthermore, we updated the network of the FAE, as it was the least sophisticated network used in the general analysis. We introduced residual connections and a second convolution at each stage to increase the depth of the originally shallow network. This configuration uses the features of the pretrained EfficientNet-B4, and we call it FAE-v2.
The specific combination of the structural similarity index (SSIM) with deep features used by the FAE showed remarkable consistency throughout the different networks employed. The FAE-v2 showed no improvements over the FAE. We argue that this observation points to fundamental problems in the design of UAD methods. A better performance on the pretext task usually does not correlate with downstream detection performance. This behavior is observable for two of the reconstruction-based approaches and for all feature-based approaches in the conducted analysis. In the main paper, we approach this problem by using the checkpoints that achieved maximum performance on the validation dataset to determine when the method is most valuable for the downstream task.

\subsection{Disyre Ablation Study}
\label{sec:Disyre}
\begin{figure*}[ht]
    \centering
    \includegraphics[scale=0.6]{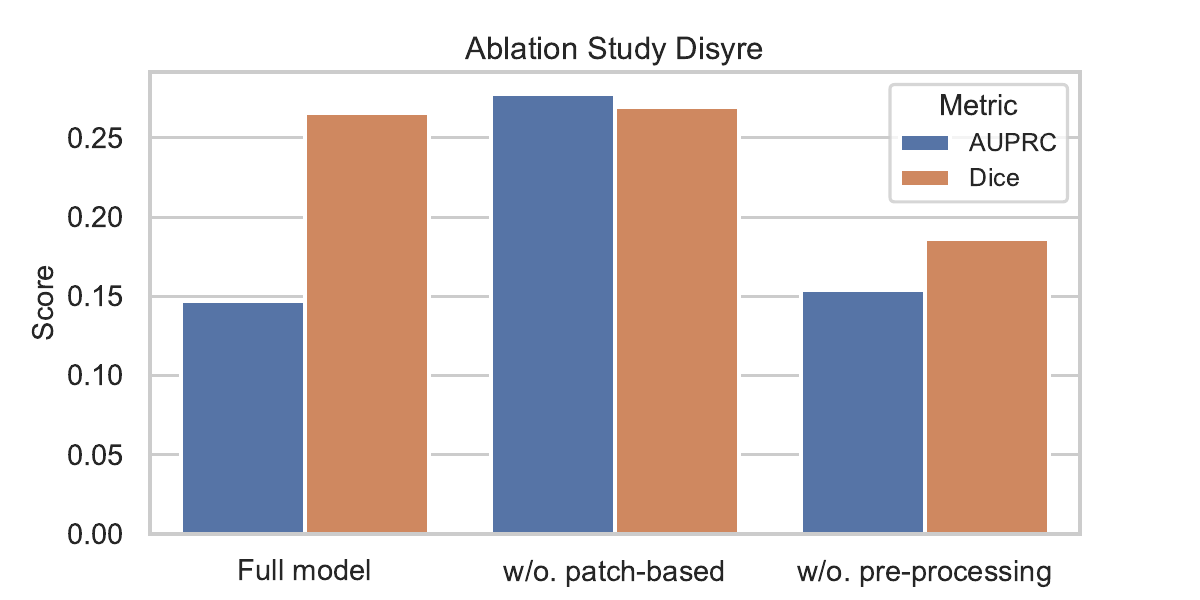}
    \caption{\textbf{Disyre Ablation Study.} Three different variants of the Disyre method have been evaluated on the validation dataset. First the full model, a model without the patch-based paradigm that takes as input the full slice and a model that omits the Disyre specific preprocessing. We observed that the specific preprocessing used is important for the Disyre model to achieve its optimum performance.}
    \label{fig:ablation}
\end{figure*}

In order to understand the outstanding performance of Disyre in our benchmark, we conducted a small ablation study on the validation dataset. This analysis is similar to the scaling analysis shown in the main part of the manuscript. We decided to remove components of the Disyre method that could be used for the majority of the other methods. These include the specific preprocessing used and the patch-based paradigm. The results can be seen in \cref{fig:ablation}. We observed that the patch-based paradigm could be removed without any performance loss; instead, we observed a small increase in performance. In contrast, the specific preprocessing used seems to be an integral part of the Disyre model and its ability to achieve state-of-the-art performance. The preprocessing includes an elastic transformation, various intensity transformations, and a mirror transformation. We did not study the effect of this preprocessing on the other methods and want to note that an interaction between the synthetic anomaly generation pipeline and this preprocessing is possible, i.e., only in combination is a robust way of generating synthetic anomalies possible.

\subsection{Model and Training Details}
Here, we report the architectures used and method-related details. In general, we aimed to reduce heterogeneity between the methods regarding preprocessing and architectural choices while maximizing individual performances. Nevertheless, a risk remains that the methods have not achieved optimal performance due to suboptimal choices of hyperparameters or misalignment between performance on the validation dataset and the test sets. All \textit{reconstruction-based} methods that originally used a U-Net-like architecture were equipped with the same U-Net, which is a slightly modified version of the DDPM++ version used in the diffusion model literature \cite{song2020score}. We opted for this choice because it ensures that older methods start on equal footing with the newly published methods. The DDPM++ model integrates the BigGAN residual blocks and scales the residual connections by $\frac{1}{\sqrt{2}}$. Note that we usually omit the scaling for all non-diffusion model methods. Additionally, we introduced a modification known as efficient U-Net, which swaps the order of the downsampling operation and the first convolution of each block \cite{saharia2022photorealistic}, use the dropout modification proposed in \cite{hoogeboom2023simple}, employ the memory efficient attention mechanism provided by PyTorch, and zero-initialize the last convolution in each block.

\subsubsection{RIAD}
In general, we build our RIAD implementation on the publicly available third-party code provided by \url{https://github.com/plutoyuxie/Reconstruction-by-inpainting-for-visual-anomaly-detection}. We removed the median filtering in the gradient magnitude similarity, as we could not find this calculation step in the original paper. Additionally, we added a small number to the square root calculation used in the edge filter to reduce the numerical problems encountered during optimization. We still experienced unstable optimization for this method throughout all the experiments. Additionally, we experimented with the masking hyperparameter $k$ to ensure coverage of large brain lesions while proportionally increasing the hyperparameter $n$, thereby maintaining sufficient contextual information for accurate reconstruction. The original implementation used $n=3$ and $k=[2,4,8,16]$. All hyperparameters can be found in \cref{Tab:RIAD}.

\begin{table}[t]
    \centering
    \caption{RIAD Hyperparameters}
    \label{Tab:RIAD}
    \begin{tabular}{c c}
    \toprule
     \textbf{Hyperparameter} &\textbf{Value} \\
     \midrule
     $\alpha$ & 1\\
     $\beta$ & 1\\
     $\gamma$ & 1\\
     k & [8,16,28,32]\\
     n & 5\\
     Update Steps & 100,000\\
     Batch size & 32\\
     Optimizer & AdamW \\
     $\beta_1, \beta_2$ & 0.9, 0.999\\
     Weight Decay & 0.01 \\
     lr & $5e-6$\\
     Image Size & 224x224\\
     Number of Input Channels & 1\\
     Model Base Dimension (Channels) & 32\\
     Channel multiplier per Resolution & (1, 1, 1, 2, 3, 4)\\
     Number of Blocks per Resolution & (1, 1, 1, 2, 3, 2)\\
     Nonlinearity & Swish \\
     Normalization & GroupNorm\\
     Attention Resolution & (14, 7)\\
     Attention Embedding Dimension & 32\\
     Dropout & 0.1 \\
     Dropout Start Resolution & 14\\
     Rescale residual connections & False\\
     Final checkpoint (number of update steps) & 100,000 
     \end{tabular}
\end{table}

\subsubsection{IterMask}
The IterMask implementation is built on the officially available implementation (\url{https://github.com/ZiyunLiang/IterMask2}). For this method, we scaled the image slices that are in the range [0,1] with $\text{slice} \cdot 6 - 3$ to bring the slices to [-3, 3], resulting in a maximum intensity number of 3 in each brain. Note that min/max normalization has been applied to the complete brain before this step. The Gaussian mask generation is performed on the complete 240 x 240 slice and then center cropped along with the rest of the image slice to 224 x 224. For mask generation and frequency masking, we keep all hyperparameters of the official implementation. All hyperparameters for both models are found in \cref{Tab:Iter}.

\begin{table}[t]
    \centering
    \caption{IterMask Hyperparameters}
    \label{Tab:Iter}
    \begin{tabular}{c c c}
    \toprule
     \textbf{Hyperparameter} &\textbf{First Model} &\textbf{Second Model} \\
     \midrule
     Update Steps & 100,000 & 100,000\\
     Batch size & 128 & 128\\
     Optimizer & AdamW & AdamW \\
     $\beta_1, \beta_2$ & 0.9, 0.999 & 0.9, 0.999\\
     Weight Decay & 0.01 & 0.01 \\
     lr & $1e-4$ & $1e-4$\\
     EMA rate & 0.9999 & 0.9999\\
     Image Size & 224x224 & 224x224\\
     Number of Input Channels & 1 & 2\\
     Model Base Dimension (Channels) & 32 & 32\\
     Channel multiplier per Resolution & (1, 1, 1, 2, 3, 4) & (1, 1, 1, 2, 3, 4)\\
     Number of Blocks per Resolution & (1, 1, 1, 2, 4, 2) & (1, 1, 1, 2, 4, 2)\\
     Nonlinearity & Swish & Swish \\
     Normalization & BatchNorm & BatchNorm\\
     Attention Resolution & (14, 7) & (14, 7)\\
     Attention Embedding Dimension & 32 & 32\\
     Dropout & 0.1 & 0.1 \\
     Dropout Start Resolution & 14 & 14\\
     Rescale residual connections & False & False\\
     Final checkpoint (number of update steps) & 100,000 & 100,000 
     \end{tabular}
\end{table}

\subsubsection{ANDi}
ANDi was developed by the first author of this paper, and this newly provided implementation can be seen as the up-to-date variant of the method. We mainly followed the diffusion model implementation of \cite{hoogeboom2023simple, salimans2022progressive} and used a Variational Diffusion Model (VDM) with a cosine schedule that is adjusted for the image size used, as in \cite{hoogeboom2023simple}. We used 56x56 as the base size for the shifting operation. Additionally, we use velocity prediction during training and learn to minimize the distance between the ground truth noise and the predicted noise, which is calculated from the velocity. During inference, we calculate the difference between the ground truth latent and the predicted latent, as in the original ANDi implementation. We aggregate the individual deviations using the geometric mean and change the Pyramid noise to standard Gaussian noise. The new noise schedule requires different values for the parameters $T_l, T_u$ and we have observed improvements in anomaly detection performance upon introducing this change. For the Pyramid noise, we opt for $c=0.9$ and image slices are scaled to be in the range of [-1, 1] before using the network or adding the noise. All hyperparameters can be found in \cref{Tab:ANDi}.

\begin{table}[t]
    \centering
    \caption{ANDi Hyperparameters}
    \label{Tab:ANDi}
    \begin{tabular}{c c}
    \toprule
     \textbf{Hyperparameter} &\textbf{Value} \\
     \midrule
     Number of Latent Variables & 1000\\
     Log SNR Minimum & -15\\
     Log SNR Maximum & 15\\
     V-Prediction & True\\
     Shift Schedule & True\\
     Time Embedding Type & Positional\\
     Pyramid Discount & 0.9\\
     $T_l$ & 25\\
     $T_u$ & 125\\
     Update Steps & 100,000\\
     Batch size & 128\\
     Optimizer & AdamW \\
     $\beta_1, \beta_2$ & 0.9, 0.999\\
     Weight Decay & 0.01 \\
     lr & $1e-4$\\
     EMA rate & 0.9999\\
     Image Size & 224x224\\
     Number of Input Channels & 1\\
     Model Base Dimension (Channels) & 32\\
     Channel multiplier per Resolution & (1, 1, 1, 2, 3, 4)\\
     Number of Blocks per Resolution & (1, 1, 1, 2, 4, 2)\\
     Nonlinearity & Swish \\
     Normalization & GroupNorm\\
     Attention Resolution & (14, 7)\\
     Attention Embedding Dimension & 32\\
     Dropout & 0.1 \\
     Dropout Start Resolution & 14\\
     Rescale residual connections & True\\
     Final checkpoint (number of update steps) & 30,000
     \end{tabular}
\end{table}

\subsubsection{Disyre}
We built our Disyre implementation on the official implementation (\url{https://github.com/snavalm/disyre}) and used Disyre v2 for all experiments. We downloaded the shapes provided on GitHub (needed for the DAG) and kept all hyperparameters of the official implementation, besides the tissue classes estimated with k-means and the architecture of the network. For the tissue classes, we run k-means on our training dataset for T1w and T2w images separately and use the respective clusters for the different training runs. As mentioned in \cref{sec:Disyre}, we maintain the specific preprocessing pipeline of Disyre as well as the patch-based paradigm for our implementation and provide results for ablations regarding these options. After the DAG pipeline, we bring the slices to [-1, 1] to follow the typical diffusion model literature. All  hyperparameters can be found in \cref{Tab:Disyre}.

\begin{table}[t]
    \caption{Disyre Hyperparameters}
    \label{Tab:Disyre}
    \begin{tabular}{c c}
    \toprule
     \textbf{Hyperparameter} &\textbf{Value} \\
     \midrule
     Number of Latent Variables & 100\\
     Beta Minimum & 0.1\\
     Beta Maximum & 20\\
     Time Embedding Type & Positional\\
     Anomaly Type & DAG\\
     Anomaly Patch Size & 64x64\\
     No Anomaly in Background & True\\
     T1w clusters & [0.6079509, 0.07953554, 0.4649098, 0.78062236, 0.31114596]\\
     T2w clusters & [0.04920235, 0.5074527, 0.2573119, 0.7149557, 0.37794548]\\
     Update Steps & 100,000\\
     Batch size & 64\\
     Optimizer & AdamW \\
     $\beta_1, \beta_2$ & 0.9, 0.999\\
     Weight Decay & 0.01 \\
     lr & $1e-4$\\
     EMA rate & 0.9999\\
     Image Size (Patch Size) & 128x128\\
     Number of Input Channels & 1\\
     Model Base Dimension (Channels) & 64\\
     Channel multiplier per Resolution & (1, 2, 2, 4, 4)\\
     Number of Blocks per Resolution & (1, 1, 2, 4, 2)\\
     Nonlinearity & Swish \\
     Normalization & GroupNorm\\
     Attention Resolution & (16, 8)\\
     Attention Embedding Dimension & 64\\
     Dropout & 0.1 \\
     Dropout Start Resolution & 8\\
     Rescale residual connections & True\\
     Final checkpoint (number of update steps) & 100,000
     \end{tabular}
\end{table}

\subsubsection{FAE}
For FAE, we used the implementation provided by the UPD study \cite{lagogiannis2023unsupervised} (\url{https://github.com/iolag/UPD_study/}), as the authors of this benchmark are also the authors of the original article. For training, we use the SSIM with a window size of five, whereas for testing, we use a window size of 11. We experienced better results with this setup on the larger images (feature maps) and the architecture of the network is adapted to work with this larger resolution. We observed no difference when using the ImageNet specific feature normalization. All hyperparameters can be found in \cref{Tab:FAE}. 

\begin{table}[t]
    \centering
    \caption{FAE Hyperparameters}
    \label{Tab:FAE}
    \begin{tabular}{c c}
    \toprule
     \textbf{Hyperparameter} &\textbf{Value} \\
     \midrule
     Update Steps & 40,000\\
     Batch size & 128\\
     Optimizer & AdamW \\
     $\beta_1, \beta_2$ & 0.9, 0.999\\
     Weight Decay & 0.01 \\
     lr & $1e-4$\\
     Image Size & 224x224\\
     Model Channels & [100, 150, 200, 300]\\
     Dropout & 0.1\\
     Pretrained Network & ResNet18\\
     Feature Map Size & 56x56\\
     Extracted Layers & ['maxpool', 'layer1', 'layer2'] \\
     Final checkpoint T1w (number of update steps) & 20,000\\
     Final checkpoint T2w (number of update steps) & 40,000\\
     \end{tabular}
\end{table}

\subsubsection{UniAD}
We built our implementation of UniAD on the official code available at \url{https://github.com/zhiyuanyou/UniAD}. We have tried different variants of UniAD by altering the feature size and masking strategy used in the modified attention mechanism but ultimately found the original hyperparameters to work best. In our experiments, it was crucial not to take many gradient update steps and not to use feature normalization. The hyperparameters can be found in \cref{Tab:UniAD}. 
\begin{table}[t]
    \caption{UniAD Hyperparameters}
    \hspace*{1cm}
    \label{Tab:UniAD}
    \begin{tabular}{c c}
    \toprule
     \textbf{Hyperparameter} &\textbf{Value} \\
     \midrule
     Update Steps & 100,000\\
     Batch size & 128\\
     Optimizer & AdamW \\
     $\beta_1, \beta_2$ & 0.9, 0.999\\
     Weight Decay & 0.01 \\
     lr & $1e-4$\\
     lr scheduler & Linear\\
     Clip Gradient Norm & 0.1\\
     Image Size & 224x224\\
     Pretrained Network & EfficientNet-B4\\
     Feature Map Size & 14x14\\
     Extracted Layers & ['features.1', 'features.2', 'features.3', 'features.4'] \\
     Position Embedding & Learned\\
     Model Hidden Dimension & 256\\
     Model Number of Heads & 8\\
     Number Encoder Layers & 4\\
     Number Decoder Layers & 4\\
     Feedforward Dimension & 1024\\
     Dropout & 0.1\\
     Nonlinearity & ReLu \\
     Feature Jitter & True\\
     Feature Jitter Scale & 20.0\\
     Jitter Probability & 1.0\\
     Neighbor Size & [7, 7]\\
     Neighbor Mask & [True, True, True]\\
     Final checkpoint (number of update steps) & 25,000\\
     \end{tabular}
\end{table}

\subsubsection{Reverse Distillation}
The RD implementation is built on the official code provided by \url{https://github.com/hq-deng/RD4AD}. We adapted the code to work with the image resolution used in the experiments and noticed that feature normalization and training for too long hurt the method's performance drastically. Hyperparameters can be found in \cref{Tab:RD}.

\begin{table}[t]
    \centering
    \caption{RD Hyperparameters}
    \label{Tab:RD}
    \begin{tabular}{c c}
    \toprule
     \textbf{Hyperparameter} &\textbf{Value} \\
     \midrule
     Update Steps & 25,000\\
     Batch size & 128\\
     Optimizer & AdamW \\
     $\beta_1, \beta_2$ & 0.9, 0.999\\
     Weight Decay & 0.01 \\
     lr & $1e-4$\\
     Image Size & 224x224\\
     Pretrained Network & ResNet18\\
     Extracted Layers & ['layer1', 'layer2', 'layer3'] \\
     Final checkpoint T1w (number of update steps) & 15,000\\
     Final checkpoint T2w (number of update steps) & 20,000\\
     \end{tabular}
\end{table}

\subsubsection{PatchCore}
The implementation of PatchCore follows the official code that can be found at \url{https://github.com/amazon-science/patchcore-inspection}. PatchCore is the only method that does not require any training and uses a memory bank instead. We randomly sampled 92 volumes from our training dataset to build the memory bank. Note that increasing the number of volumes would drastically increase the time required to build the memory bank. When using the 92 volumes and the greedy coreset subsampling to reduce the size to 1\%, the time to build the bank is roughly three days. We experimented with different patch sizes, strides, and memory bank sizes but found the setup working for MVTec anomaly localization to be well suited for our framework as well. Slight improvements were observed when the number of neighbors to search for in the memory bank was set to one. Hyperparameters can be found in \cref{Tab:PatchCore}. 

\begin{table}[t]
    \centering
    \caption{PatchCore Hyperparameters}
    \label{Tab:PatchCore}
    \begin{tabular}{c c}
    \toprule
     \textbf{Hyperparameter} &\textbf{Value} \\
     \midrule
     Image Size & 224x224\\
     Pretrained Network & WideResNet50\\
     Extracted Layers & ['layer2', 'layer3'] \\
     Number of Neighbors & 1\\
     Patch Size & 5\\
     Patch Stride & 1\\
     Target Embedding Dimension & 1024\\
     Sample Size & 0.01
     \end{tabular}
\end{table}